\RequirePackage{fix-cm}
\documentclass[twocolumn]{svjour3}          
\smartqed  

\usepackage{cite}
\usepackage[pdftex]{graphicx}
\usepackage{ragged2e}
\usepackage[tight,footnotesize]{subfigure}
\usepackage{rotating}
\usepackage{graphicx}
\usepackage{amsmath,amssymb} 
\usepackage{array}
\usepackage{multirow}
\usepackage{colortbl}
\usepackage{amsfonts}
\usepackage{pifont}
\usepackage{xspace}
\usepackage{etoolbox}
\usepackage{overpic}
\usepackage{color}
\usepackage{xcolor}
\usepackage{microtype}
\usepackage{xurl}
\definecolor{hfxpink}{rgb}{0.99, 0.91, 0.95}
\definecolor{mygreen}{rgb}{0, 0.6, 0}

\usepackage{esvect}		
\usepackage{algorithm}
\usepackage{algorithmic}
\usepackage[breaklinks=true]{hyperref}
\usepackage{graphicx}
\usepackage{wrapfig}
\usepackage{ragged2e}

\newcommand{\orcid}[1]{\href{https://orcid.org/#1}{\includegraphics[width=10pt]{ocrid.png}}}

\def\etal{{\em et al}}

\usepackage{hyperref}
\hypersetup{breaklinks=true,citecolor=blue, colorlinks}

\graphicspath{{./Imgs/}}
\DeclareGraphicsExtensions{.pdf,.jpg,.png}

\usepackage{silence}
\journalname{Research Article}

\begin{document}
\title{Unsupervised Robust Domain Adaptation: Paradigm, Theory and Algorithm}


\author{Fuxiang Huang\textsuperscript{1, 3}\and
        Xiaowei Fu\textsuperscript{1, 2} \and 
        Shiyu Ye\textsuperscript{2} \and
        Lina Ma\textsuperscript{2}\and
        Wen Li\textsuperscript{4}\and
        Xinbo Gao\textsuperscript{5}\and
        David Zhang\textsuperscript{6}\and
        Lei Zhang\textsuperscript{1, 2*}
}

\authorrunning{F. Huang \etal}

\institute{
\textsuperscript{1} Chongqing Key Laboratory of Bio-perception and Multimodal Intelligent Information Processing, Chongqing University,  Chongqing 400044, China \\
\textsuperscript{2} School of Microelectronics and Communication Engineering,  Chongqing University, Chongqing 400044, China \\
\textsuperscript{3} School of Data Science, Lingnan University Hong Kong, Hong Kong, China \\
\textsuperscript{4} School of Computer Science and Engineering,  University of Electronic Science and Technology of China,  Chengdu 611731,  China.\\
\textsuperscript{5} Chongqing Key Laboratory of Image Cognition,  Chongqing University of Posts and Telecommunications,  Chongqing 400065,  China.\\
\textsuperscript{6} School of Science and Engineering,  Chinese University of Hong Kong (Shenzhen),  Shenzhen 518172,  China.
\\
\textsuperscript{*}Corresponding author: Lei Zhang (leizhang@cqu.edu.cn)
}

\date{Received: date / Accepted: date}
\renewcommand{\makeheadbox}{} 
\maketitle








\begin{abstract}
{
Unsupervised domain adaptation (UDA) aims to transfer knowledge from a label-rich source domain to an unlabeled target domain by addressing domain shifts. Most UDA approaches emphasize transfer ability, but often overlook robustness against adversarial attacks. Although vanilla adversarial training (VAT) improves the robustness of deep neural networks,  it has little effect on UDA. This paper focuses on answering three key questions: 1) Why does VAT,  known for its defensive effectiveness,  fail in the UDA paradigm? 2) What is the generalization bound theory under attacks and how does it evolve from classical UDA theory? 3) How can we implement a robustification training procedure without complex modifications? Specifically,  we explore and reveal the inherent \textit{entanglement challenge} in general UDA+VAT paradigm,  and propose an unsupervised robust domain adaptation (URDA) paradigm. We further derive the generalization bound theory of the URDA paradigm so that it can resist adversarial noise and domain shift. To the best of our knowledge,  this is the first time to establish the URDA paradigm and theory.
We further introduce a simple,  novel yet effective URDA algorithm called Disentangled Adversarial Robustness Training (DART),  a two-step training procedure that ensures both transferability and robustness. DART first pre-trains an arbitrary UDA model,  and then applies an instantaneous robustification post-training step via disentangled distillation. 
Experiments on four benchmark datasets with/without attacks show that DART effectively enhances robustness while maintaining domain adaptability,  and validate the URDA paradigm and theory.}

 
\end{abstract}

\section{Introduction}\label{sec1}

Unsupervised domain adaptation (UDA)~\cite{wilson2020survey, fang2024source} can effectively learn a classification model by transferring relevant and useful knowledge from labeled source domain to unlabeled target domain. Recently,  domain adaptation methods combined with deep neural networks (DNNs) have achieved a significant success~\cite{ganin2015unsupervised, long2015learning, long2017deep, long2018conditional, saito2018maximum, li2019joint, li2020deep,  zhang2019bridging, pan2019transferrable, cui2020towards,  li2020enhanced, xu2021cdtrans,  na2021fixbi,  li2021bi, xiao2021dynamic,  wei2021metaalign,  gao2021gradient,  Zhu_2023_CVPR}. However,  the predominant focus of these approaches is the transferability of UDA models. Although DNN robustness is widely recognized~\cite{zhang2019theoretically, wang2019improving, chen2023egans, chen2022transzero++, chen2025semantics},  UDA robustness is surprisingly paid little attention. We resort to clarify \textit{if DNN and UDA are the same thing in improving robustness and if the well-known adversarial training for DNN can be transplanted to UDA}. Unfortunately,  their findings in DNN (single domain) do not generalize well to UDA models (two domains). To differentiate from the non-robust UDA paradigm,  in this paper,  we call the proposed robust paradigm as \textbf{unsupervised robust domain adaptation (URDA)}.

\begin{figure}[h]
	\centering
    \includegraphics[width=0.5\textwidth]{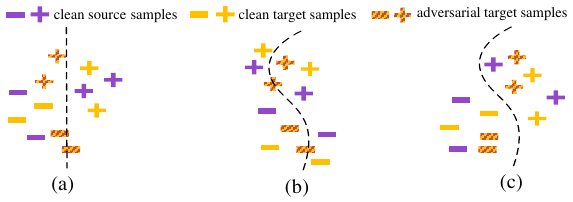}
	\caption{Schematic result of standard UDA paradigm,  UDA+VAT paradigm,  and the proposed URDA paradigm. (a) Standard UDA model cannot classify adversarial samples. (b) Robust UDA model with adversarial training can classify adversarial samples better but leads to misclassification of clean samples. (c) The proposed DART can classify both clean and adversarial samples well.}
	\label{fig:class}
\end{figure}

In recent years,  the rise of adversarial attack methods~\cite{szegedy2013intriguing, goodfellow2014explaining, kurakin2018adversarial, madry2017towards, wiyatno2018maximal,  carlini2017towards,  rony2019decoupling,  wu2020stronger,  xiao2018spatially,  eykholt2018robust,  croce2020reliable,  augustin2020adversarial},  enforces researchers aware of the fragility of DNNs. However,  investigations into model robustness predominantly focus on general DNN models,  such as adversarial defense~\cite{zhang2019theoretically, wang2019improving,  goldblum2020adversarially,  zhang2020robustified,  yang2021exploring, Zhang2024tpamiUAE, Li_2024_neurips, Kim_2023_CVPR, Yue_2024_CVPR, Yin_2024_CVPR} and adversarial distillation~\cite{Jung_2024_CVPR, Li_2024_CVPR, Huang_2023_CVPR, Liu_2023_CVPR, zhu2021reliable, zi2021revisiting},  while less attention is paid to UDA models~\cite{li2023pseudo, zhang2020robustified,  2020Adversarially,  yang2021exploring,  Lo_2022_ACCV, Gao_2023_ICCV}. Notably it is extremely challenging to train a robust UDA model without sacrificing the classification accuracy of clean (benign)  target samples. Fig.~\ref{fig:class} (a) represents the traditional UDA method and numerous misclassifications of adversarial target samples are witnessed,  which indicates the vulnerability of UDA. Fig.~\ref{fig:class} (b) depicts the robust UDA methods obtained using the vanilla adversarial training (VAT),  which not only improves little on adversarial robustness but drops clean accuracy. The in-depth insights why UDA+VAT paradigm does not commonly work are formally discussed,  explored and resolved in this paper. Fig.~\ref{fig:class} (c) shows the objective of our proposed URDA paradigm,  a new robustification approach. After delving deeper into the combined paradigm of UDA (transfer training)+VAT (adversarial training),  we first unveil the inherent \textit{entanglement} problem between transfer training and adversarial training. The idea of the proposed URDA paradigm is also originated from this finding of inherent entanglement. 

\begin{figure}[h]
	\centering
    \includegraphics[width=0.5\textwidth]{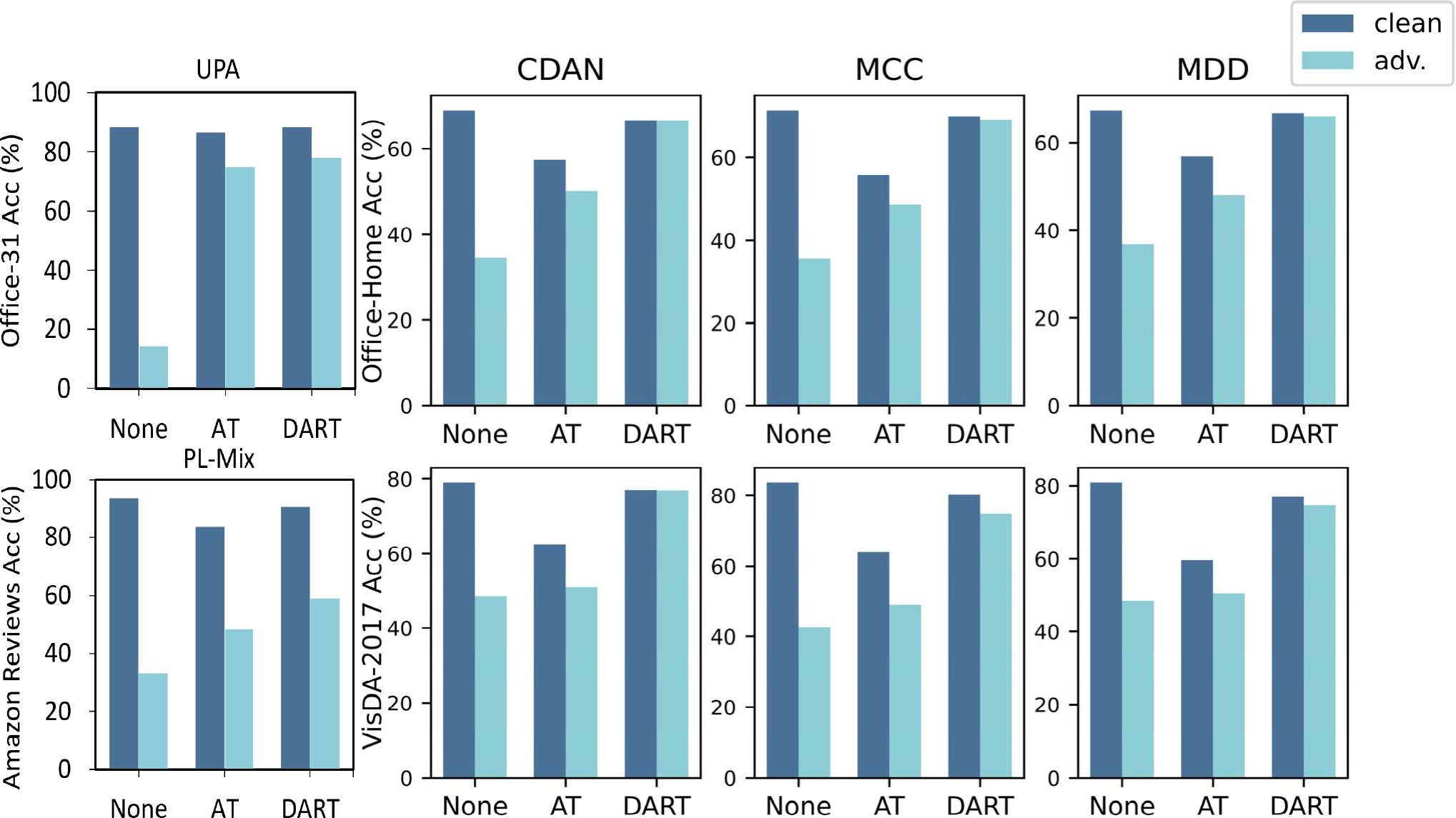}
	\caption{Pilot exploratory experiments on the robustness analysis of vanilla UDA models (Pseudo-labeling based methods: UPA~\cite{chen2024uncertainty} and PL-Mix~\cite{kong2024unsupervised}, traditional CDAN~\cite{long2018conditional},  MCC~\cite{DBLP:journals/corr/abs-1912-03699} and MDD~\cite{zhang2019bridging}), effectiveness analysis of vanilla adversarial training (AT) for robust UDA,  and superiority analysis of the proposed URDA paradigm (i.e., DART). The clean target accuracy and adversarial target accuracy (adv.) are reported on Office-31, Amazon Reviews, Office-Home and VisDA-2017 benchmarks. The vulnerability of UDA models is shown, while the proposed DART well improves their robustness.}
	\label{fig:pilot}
\end{figure}

\begin{figure*}[h]
	\centering
	\subfigure{
		\includegraphics[width=2.1in]{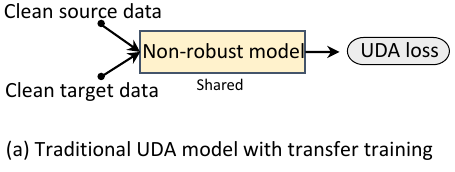}}
	\subfigure{
		\includegraphics[width=2.1in]{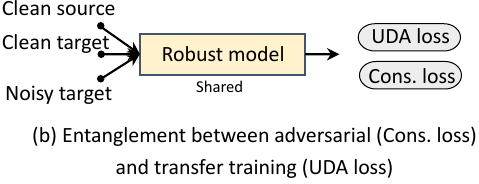}}
	\subfigure{
		\includegraphics[width=2.1in]{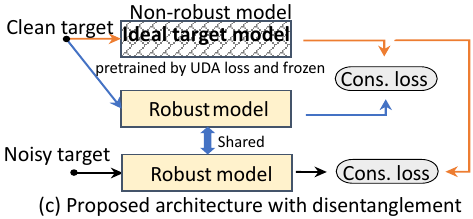}}
	\caption{Architecture for standard UDA non-robust paradigm,  UDA+VAT robust paradigm and the proposed URDA robust paradigm,  resp. (a) Traditional UDA model. (b) Robust UDA+vanilla adversarial training (VAT) by directly imposing consistency constraints between clean and adversarial target samples (entanglement). (c) The proposed URDA paradigm by disentangling transfer training from adversarial training,  where Cons. means consistency constraint. In essence,  URDA aims to robustify a pretrained UDA model by a robust model supervised with Cons. loss.}
	\label{fig:entang} 
\end{figure*}
To further confirm our hypothesis,  we conduct pilot experiments to explore the robustness of UDA to adversarial attacks and the feasibility of VAT. The classification accuracy of clean target samples and adversarial target samples are shown in Fig.~\ref{fig:pilot},  which is manifested on five representative UDA models (Pseudo-labeling based methods: UPA~\cite{chen2024uncertainty} and PL-Mix~\cite{kong2024unsupervised}, traditional CDAN~\cite{long2018conditional},  MCC~\cite{DBLP:journals/corr/abs-1912-03699} and MDD~\cite{zhang2019bridging}) and four benchmark datasets (Office-31~\cite{venkateswara2017deep}, Amazon Reviews~\cite{saenko2010adapting}, Office-Home~\cite{venkateswara2017deep} and VisDA-2017~\cite{peng2017visda}). Note that UDA generally focuses on the target domain accuracy. Therefore,  in this exploratory experiment,  we perform the I-FGSM~\cite{kurakin2018adversarial} based AT on the target domain and generate adversarial samples for evaluation. I-FGSM performs multiple small-step updates in the direction of the gradient sign, projecting the adversarial example back into the allowed perturbation range at each iteration. During training, VAT forces the model to make the same predictions for clean and adversarial examples. As is shown in Fig.~\ref{fig:pilot},  we have two observations: \textbf{1)} For each UDA model,  the target domain performance degrades seriously while encountering adversarial samples. This indicates that adversarial samples have a devastating impact on domain adaptation ability. \textbf{2)} For VAT,  although it improves a little on the adversarial target samples,  it does not show relatively optimistic defensive ability as it appears in single-domain image classification task,  and also the clean accuracy drops too much. It is easy to understand the first observation because it has similar vulnerability with conventional deep networks as adversarial attacks appear. But for the second observation,  it is surprised that VAT,  known for its defensive effectiveness,  fails to transplant for protecting UDA against attacks. So,  \textit{we cannot help but ask why this is happening?} Next,  let's have a discussion.

Two natural questions are arise. 1) \textit{Why this phenomenon happens?} 2) \textit{How to achieve a model,  exemplified in Fig.~\ref{fig:class} (c),  that not only ensures the classification accuracy of clean target samples (transferability) but also improves the accuracy of adversarial samples (robustness)?} 3) \textit{What is the generalization bound of URDA theory?} We resort to answer these questions by revealing a crucial factor impacting the accuracy of clean samples: \textit{the consistency constraint between clean and adversarial target samples from a shared network in UDA+VAT paradigm}.  Traditional AT applied on unlabeled target domain for robustness may inadvertently compromise transfer training (i.e.,  UDA) and leads to model collapse,  because it may mislead the model to predict clean samples as incorrect classes only by simply constraining consistency between clean and adversarial target samples. Consequently,  the entire training process becomes extremely unstable and even converges towards a detrimental direction. To support this claim, we conduct a pilot exploratory experiment for several typical UDA models in Fig.~\ref{fig:pilot}. We observe that the accuracy of clean samples with vanilla adversarial training (i.e., AT) degrades too much, while the accuracy of adversarial samples improves little. This clearly illustrates the instability caused by the direct consistency constraint. To solve that, we propose DART, which eliminates this direct connection to disentangle transfer learning from adversarial training, and achieves significant trade-off between accuracy and robustness. The conventional UDA paradigm (i.e., transfer training) is formulated as Fig.~\ref{fig:entang} (a). For simplicity,  we describe the above phenomenon as \textit{interactive entanglement} between adversarial training and transfer training as is shown in Fig.~\ref{fig:entang} (b),  in which the clean data consist of labeled source data and unlabeled target data,  while the noisy data means the adversarial samples of unlabeled target data.

The \textit{entanglement} essentially arises from a conflict between the adversarial training and transfer training on unlabeled target domain. \textit{Trade-offing the clean and noisy target data without label supervision is a challenge,  which easily leads to mode collapse due to explosive error accumulation on target.} To address the challenge,  we introduce an ideal target classifier to replace the traditional consistency constraint between adversarial and clean target samples,  decoupling from a generally designed shared network,  as is shown in Fig.~\ref{fig:entang} (c). Specifically,  we deduce the URDA generalization bound for expected target error under adversarial noise to ensure the robustness,  which makes a fundamental progress of UDA. The ideal target model is trained off-line by an arbitrary vanilla UDA model using labeled source and unlabeled target domains,  and then is frozen. This new upper bound removes direct constraints between clean and adversarial samples,  theoretically achieves robust UDA training and overcomes malicious entanglement challenge,  which is further detailed in following section.

\begin{figure*}[h]
	\centering
	\includegraphics[width=1\textwidth]{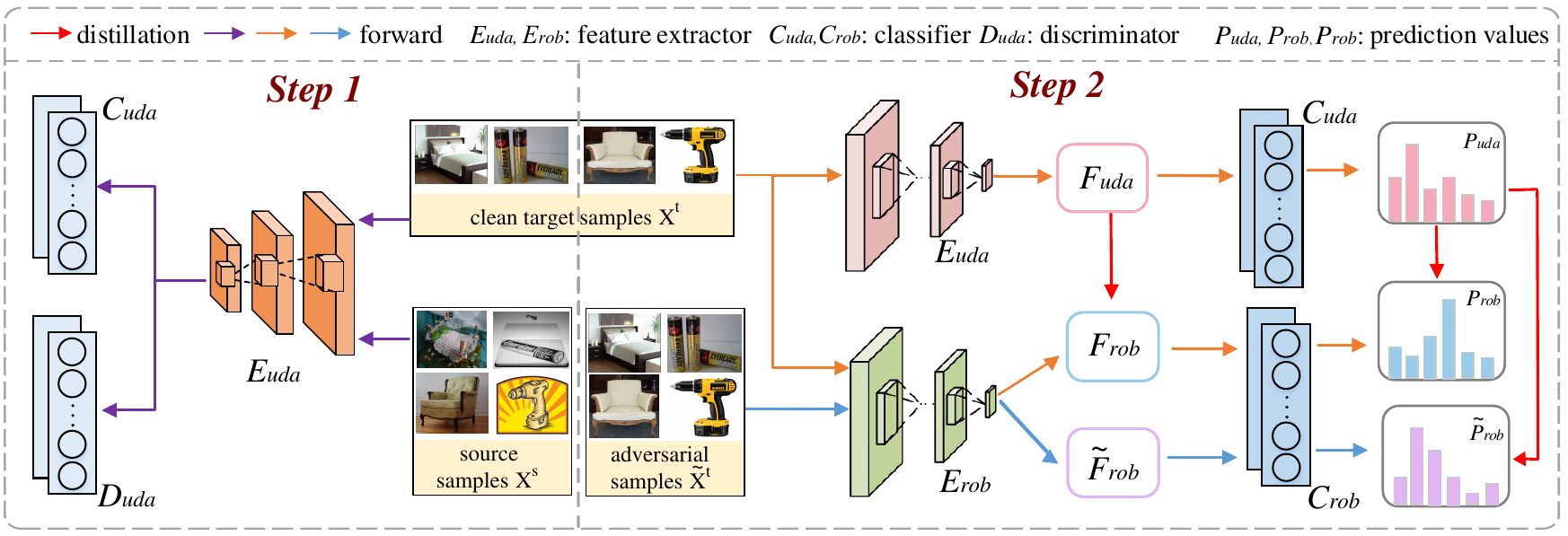}
	\caption{The pipeline of the proposed DART, derived from the proposed URDA theoretical bound under adversarial noises, is composed of two training steps: pre-training step and post-training step. In \emph{Step 1},  a non-robust model $M_{uda}=E_{uda}\cup{C_{uda}}$ is pre-trained by an arbitrary UDA method,  and frozen. In \emph{Step 2},  we employ the pre-trained non-robust UDA model $M_{uda}$ to guide the on-the-fly robustification post-training of the robust  model $M_{rob}=E_{rob}\cup{C_{rob}}$ via disentangled distillation.} 
\label{fig:2}
\end{figure*}

On the basis of the URDA theory,  we further propose a simple yet effective URDA algorithm,  i.e.,  Disentangled Adversarial Robustness Training (DART),  which can be viewed as the first URDA baseline. Technically,  DART consists of two disentangled training steps. In \textbf{step 1},  a classical UDA model is pre-trained and frozen as the ideal target classifier. In \textbf{step 2},  by frozen the pre-trained UDA model,  we perform robustification post-training on the fly via knowledge distillation. This approach significantly enhances model's robustness to adversarial samples, while preserving accuracy of clean samples. \textit{By disentangling the non-robust transfer training from the robust adversarial training,  this training paradigm effectively mitigates the entanglement dilemma between the two programs}. The main contributions and novelty are summarized as follows:

\begin{itemize}
	\item We theoretically unveil the entanglement challenge between adversarial training and transfer training in the UDA+VAT paradigm,  and establish,  to the best of our knowledge,  the first unsupervised robust domain adaptation (URDA) paradigm with a mathematically derived generalization bound theory.
	
	\item Based on the proposed URDA theory,  we propose a two-step on-the-fly URDA training algorithm,  i.e.,  DART,  by disentangling the adversarial training from transfer training. It first pre-trains an arbitrary UDA model and then performs robustification post-training step with theoretical guarantee. Fig.~\ref{fig:2} depicts the overall algorithm architecture of DART.
	
	\item The proposed DART is evaluated and verified on four benchmark datasets using three vanilla UDA models. Exhaustive experiments demonstrate that DART can considerably improve the model robustness against attacks in target domain on the premise of maintaining excellent domain transferability of clean (benign) samples.
\end{itemize}

\section{Related Work}\label{sec2}

\textbf{~~~Adversarial Attack.}
Szegedy et al. \cite{szegedy2013intriguing} pinpoint the vulnerability of deep neural networks to imperceptible noise. Recent advances in adversarial attack methods fall into three main categories: 1) \emph{gradient-based approaches},  such as FGSM~\cite{goodfellow2014explaining},  I-FGSM~\cite{kurakin2018adversarial},  PGD~\cite{madry2017towards},  and JSMA~\cite{wiyatno2018maximal}; 2) \emph{Optimization-based techniques},  such as CW~\cite{carlini2017towards} and DDN~\cite{rony2019decoupling}; 3) \emph{Space-based attacks},  such as FWA~\cite{wu2020stronger},  STA~\cite{xiao2018spatially},  and RP2~\cite{eykholt2018robust}. Additionally,  a powerful attack known as AutoAttack~\cite{croce2020reliable} combines four techniques: AutoPGD~\cite{augustin2020adversarial},  DLR~\cite{croce2020reliable},  FAB~\cite{wu2020stronger},  and Square attack~\cite{andriushchenko2020square}.

\textbf{Adversarial Training based Defense.}
We concentrate on distillation-based adversarial training defense,  categorized into self-distillation~\cite{zhang2019theoretically, wang2019improving} and teacher-based distillation~\cite{goldblum2020adversarially, zhu2021reliable, zi2021revisiting}. For example,  TRADES~\cite{zhang2019theoretically} employs soft labels for self-distillation,  improving robustness compared to traditional hard-label-based adversarial training methods. MART~\cite{wang2019improving} proposes a weighted method for calculating Kullback-Leibler loss.
For teacher network distillation,  ARD~\cite{goldblum2020adversarially} distills clean samples from the robust teacher network.RSLAD~\cite{zi2021revisiting} distills student networks from robust teacher networks by instructing both clean and adversarial samples. \emph{In contrast to these methods,  our approach does not require a sufficiently robust teacher network to provide reliable supervision for traditional UDA models. Instead,  we aim to establish a new theoretical bound and derive an easily implemented firewall training algorithm for protecting unsupervised domain adaptation against adversarial attacks.}

\textbf{Unsupervised Domain Adaptation.}
UDA with deep neural networks can be broadly categorized into two main types. 1) Information statistics-based approaches \cite{long2015learning,  long2017deep,  li2020deep,  zhang2019bridging,  pan2019transferrable,  cui2020towards,  li2020enhanced}
employ statistical discrepancy metrics to quantify domain shifts and addresses them by minimizing statistical differences. For instance,  DAN~\cite{long2015learning} utilizes multi-kernel maximum mean discrepancy (MMD) across various task-specific layers. JAN~\cite{long2017deep} integrates MMD to enforce alignment in joint domain distributions. 2) Domain adversarial learning-based approaches~\cite{ganin2015unsupervised, long2018conditional, saito2018maximum, li2019joint, li2021bi, xiao2021dynamic, wei2021metaalign} aim to learn domain-invariant feature representations through the interaction between domain classifiers and feature extractors. For example,  DANN~\cite{ganin2015unsupervised} introduces a domain discriminator to facilitate domain adaptation. CDAN)~\cite{long2018conditional} proposes a conditional domain discriminator that adjusts domain adaptation based on discriminative information,  enhancing the alignment between different distributions in the source and target domains. \cite{yu2022align} uses a two-stage framework with DCAN and PCST to resolve misalignment between representation learning and semi-supervised methods in unsupervised domain adaptation. However, the robustness of UDA is not considered.

\textbf{Adversarial Defense for UDA.}
Despite the development of a number of UDA models,  their safety and robustness are considerably overlooked. \cite{awais2021adversarial} leverages intermediate representations learned by multiple robust ImageNet models to improve the robustness of UDA models. Yang et al. \cite{yang2021exploring} propose to maximize the agreement between clean images and their adversarial examples by a contrastive loss in the output space. Based on the self-training paradigm,  SRoUDA~\cite{zhu2022srouda} trains an adversarial target model on pseudo-labeled target data and fine-tunes the source model with a meta-step. Under the guidance of high-quality pseudo labels,  \cite{10043655} proposes to align the representations of target domain clean and adversarial samples with the source anchors. \cite{soni2025toward} adopts a self-training method that enhances semantic feature alignment and pseudo-label reliability through attention consistency and adversarial refinement. However,  all these works just propose an algorithm without theoretical guarantees on the robustness of UDA models and overlook practicality. This paper addresses this problem and firstly proposes new insights from the perspectives of new paradigm, theory and algorithm.

\section{URDA: Theory and Algorithm}\label{sec3}

\subsection{Notations}\label{subsec3.1}
In URDA,  there is a labeled source domain $\left\{\left(x_{i}^{s},  y_{i}^{s}\right)\right\}_{i=1}^{m}$ drawn from the source distribution $\mathcal{D}_{S}$ on $\mathcal{X} \! \times \! \mathcal{Y}$,  where $\mathcal{X}$ is the sample set and $\mathcal{Y}$ is the label set $\{1,  \ldots,  K\}$ in multi-class classification. We denote a labeling function for source distribution $f_{S}\!: \!\mathcal{X} \!\rightarrow \!\mathcal{Y}$,  $y_{i}^{s}\!=\!f_{S}\left(x_{i}^{s}\right)$. Simultaneously,  an unlabeled target domain $\left\{\left(x_{i}^{t}\right)\right\}_{i=1}^{n}$ is drawn from the natural target distribution $\mathcal{D}_{T}$.
We denote a loss function $\mathcal{L}\!: \!\mathcal{Y} \! \times \! \mathcal{Y} \! \rightarrow \! \mathbb{R}$ defined over pairs of labels. For multi-class classification,  we denote a hypothesis function $h\!:\! \mathcal{Y} \! \times \! \mathcal{Y} \!\rightarrow \! \mathbb{R}$. For any hypothesis function $h\! \in \!\mathcal{H}$ on distribution $\mathcal{D}_S$,  we denote $\epsilon_{S}(h,  f)\!=\!\mathbb{E}_{x \sim \mathcal{D}_S} \mathcal{L}(h(x),  f(x))$ as the expected risk,  abbreviated as $\epsilon_{S}(h)$,  where $f$ is the labeling function. Similar notations apply to the target domain $\mathcal{D}_T$. 

We define an adversarial target domain sample $\tilde{x}^{t}$,  which is an adversarial sample of $x^{t}$ generated by an adversarial attacker $\beta(\cdot)$. Specifically,  
\begin{equation}
\tilde{x}^{t}\!=\!\beta(x^t)\!=\!\underset{ {\|\tilde{x}^{t}-x^{t}\|_{p} \leq \epsilon}}{\arg\max } \mathcal{L}_{c e}\left(g(\tilde{x}^t),  \hat{y}\right),  
\end{equation}
where $\beta$ follows the standard assumption in adversarial learning. Specifically, the constraint $|\tilde{x}^{t} - x^{t}|_{p} \leq \epsilon$ ensures human-imperceptible perturbations, where $\epsilon$ is the perturbation budget, $|\cdot|_p$ denotes the $\ell_p$-norm (with $\ell_\infty$ used in this work), $g$ is the UDA classifier, $\hat{y}$ is the pseudo-label of $x^t$, and $\mathcal{L}_{ce}$ represents the cross-entropy loss. We assume the perturbation is small enough to preserve the semantic label of the clean sample (perturbation locality) and that the induced adversarial distribution is stationary under a fixed worst-case perturbation budget~\cite{madry2017towards}. These assumptions are consistent with mainstream adversarial training works, which guarantee that our URDA theory holds across different attacks and budgets. We define the adversarial target domain distribution as $\mathcal{D}_{\widetilde{T}}$, in contrast to the original (clean) target distribution $\mathcal{D}_T$,  and the adversarial target domain samples $\left\{\left(\tilde{x}_{i}^{t}\right)\right\}_{i=1}^{n}$ are drawn from $\mathcal{D}_{\widetilde{T}}$.

We denote the model as ${M}$,  which  consists of a feature extractor ${E}$ and a classifier ${C}$. Besides,  ${D}$ serves as the domain discriminator and $F$ represents the feature extracted from ${E}$.

\subsection{Entanglement Challenge in UDA+VAT Paradigm}\label{subsec3.2}

For further interpreting the \textit{interactive entanglement} in the combined paradigm of UDA+VAT for robust UDA training,  i.e. Fig.~\ref{fig:entang} (b),  we show the upper bound of expected target risk for both clean and adversarial target domain samples in adversarial training by directly adding consistency constraint between them.

\begin{proposition}\label{propsition1}
According to the triangle inequality for classification error $\epsilon$~\cite{ben2006analysis},  it implies that for any labeling function $f_{1}$, $f_{2}$,  and $f_{3}$,  $\epsilon\left(f_{1},  f_{2}\right)\! \leq \! \epsilon\left(f_{1},  f_{3}\right)+ \epsilon\left(f_{2},  f_{3}\right)$. For any hypothesis $h \! \in \! \mathcal{H}$,  the upper bound of expected target risk $\epsilon_{T,  \widetilde{T}}(h)$ in adversarial training is given by
\begin{equation}
		\begin{split}
			\epsilon_{T,  \widetilde{T}}\left(h,  f_{T}\right)
			&=\epsilon_{\widetilde{T}}\left(h(\tilde{x}),  f_{T}\right)+\epsilon_{T}\left(h(x),  f_{T}\right) \\
			& \leq \epsilon_{\widetilde{T}, T}\left(h(\tilde{x}),  h(x)\right)+ \epsilon_{T}\left(h(x),  f_{T}\right) \\
			& ~~~+ \epsilon_{T}\left(h(x),  f_{T}\right) \\
			& \leq \epsilon_{\widetilde{T}, T}\left(h(\tilde{x}),  h(x)\right)+2 \epsilon_{T}\left(h(x),  f_{T}\right)
		\end{split}	
	\label{aa}
\end{equation}
\emph{where $h(\cdot)$ represents the function that classifies the perturbed samples $\tilde{x}$ generated by an adversarial attacker $\beta(\cdot)$,  i.e.,  $h(\tilde{x})=h(\beta(x))$,  and $f_T$ is the labeling function of target domain. Clearly,  $f_T$ is unknown for UDA.}
\end{proposition}

From Inequality (\ref{aa}),  the upper bound of the expected target risk $\epsilon_{T, \widetilde{T}}(h)$ is composed of two error terms. The first error term denotes the prediction discrepancy between the pair-wise adversarial and clean samples. The smaller the error is,  the more robust the model is. The second error term is the natural target error of an arbitrary UDA model,  which cannot be directly calculated in the UDA model due to the lack of target labels. But it can be estimated by the upper bound of expected target error on a classical UDA model \cite{ben2010theory}. Minimizing the second error term can only improve the classification accuracy of the model on clean target samples (the objective of the traditional UDA models),  but not robust to the adversarial target samples. Therefore,  this paper aims to reduce the first error term towards robust domain adaptation (a.k.a. defense against attacks),  while simultaneously minimizing the second term for protecting natural samples (a.k.a. benign maintenance).

If the two terms are sufficiently small,  the upper bound in Inequality (\ref{aa}) is valuable and available. However,  if one of the two terms becomes smaller and the other one becomes larger,  the model cannot effectively lower the upper bound $\epsilon_{T, \widetilde{T}}(h)$. From Inequality (\ref{aa}),  we can observe that ${h(x)}$ in the upper bound of the expected target risk is affected by both $h(\tilde{x})$ and $f_{T}$. The purpose of the perturbation function $\beta(\cdot)$ is to make $h(\tilde{x}_{i}) \neq f_{T}\left(x_{i}\right)$,  thus when we try to reduce one error term,  another error term may increase. \textit{This is coined the irreconcilable entanglement of the two terms,  which makes the two error terms difficult to be simultaneously minimized during optimization,  and thus loses adversarial robustness.} Fig.~\ref{fig:entang} (b) shows the UDA+VAT paradigm with entanglement between transfer training and adversarial training. In the following,  we introduce the URDA theory by disentangling transfer training and adversarial training,  as is shown in Fig.~\ref{fig:entang} (c).

\textit{Why does VAT fail in UDA?} Unlike standard adversarial training in the single-domain setting, VAT in UDA operates without reliable label supervision on the target domain. This makes the consistency constraint between clean and adversarial target samples prone to error amplification, as incorrect pseudo-labels may be reinforced. Furthermore, VAT directly implement transfer training (aligning clean target features with source features) and adversarial training (enhancing robustness), which, due to the entanglement challenge, often results in a conflict, i.e. improving adversarial accuracy often sacrifices clean accuracy. Our exploratory results in Fig.~\ref{fig:pilot} empirically confirm this phenomenon: VAT slightly improves adversarial robustness but leads to a significant degradation in clean accuracy. These observations emphasize that robustness cannot be simply inherited from single-domain adversarial training.

\subsection{The Proposed URDA Generalization Bound Theory}
From the above analysis,  we can conclude that removing entanglement and simultaneously minimizing the two error terms of the upper bound is critical for developing a robust UDA model. Therefore,  we propose a natural idea to disentangle the transfer training and adversarial training,  which prevents the interaction between them. More specifically, we redefine the upper bound of the expected target risk $\epsilon_{T, \widetilde{T}}({h, f_T})$ in inequality~(\ref{aa}) by replacing ${h(x)}$ for clean target samples with an ideal target classifier $h_{t}^{*}$.
\begin{definition}\label{definition1} Define an ideal target classifier:
\begin{equation}
	h_{t}^{*}=\underset{h \in \mathcal{H}}{\operatorname{argmin}} \operatorname{Pr}_{x,  y \sim \mathcal{D}_{\mathrm{T}}}(h(x) \neq y)
\label{eqtargetclassifier}
\end{equation}
where $h_{t}^{*}$ has lower classification risk in unlabeled target domain and $y$ denotes target label,  which is just used here for analysis but actually unknown.
\end{definition}

Consider that we would like to develop a firewall for arbitrary pre-trained UDA model against attacks,  to make the ideal target classifier explicit,  we can employ arbitrary vanilla UDA methods (\textbf{CDAN}~\cite{long2018conditional},  \textbf{MCC}~\cite{DBLP:journals/corr/abs-1912-03699} and \textbf{MDD}~\cite{zhang2019bridging} are explored in experiments) to obtain such an ideal target classifier with lower risk by minimizing the following upper bound of target error as follows \cite{ben2010theory}:
\begin{equation}
	\epsilon_{T}\left(h_{t}^{*},  f_{T}\right) \leq \epsilon_{S}(h)+\frac{1}{2} d_{\mathcal{H} \Delta \mathcal{H}}(S,  T)+\lambda
\label{udatheory}
\end{equation}
where $\epsilon_{S}(h)$ is the expected error on source domain,  $d_{\mathcal{H} \Delta \mathcal{H}}(S,  T)$ is $\mathcal{H} \Delta \mathcal{H}$-divergence between two domains and $\lambda$ is an ideal joint error considered sufficiently small.

With the Proposition \ref{propsition1} and Definition \ref{definition1},  we formally specify the proposed target error upper bound of URDA problem with disentanglement as the following theorem.
\begin{theorem}\label{thm2}
For any hypothesis $h \in \mathcal{H}$ and an ideal target classifier $h_t^*$,  the redefined upper bound of the expected target risk $\xi_{T, \widetilde{T}}({h, f_T})$ for both the natural target samples denoted by $x$ and the adversarial target samples denoted by $\widetilde{x}$ is formulated as:	
\end{theorem}
\begin{equation}
\small
		\begin{split}
			\xi_{T, \widetilde{T}}\left(h,  f_{T}\right)
			&=\epsilon_{\widetilde{T}}\left(h(\tilde{x}),  f_{T}\right)+\epsilon_{T}\left(h(x),  f_{T}\right) \\
			& \leq \epsilon_{\widetilde{T}}\left(h(\tilde{x}),  h_{t}^{*}({x})\right)+\epsilon_{T}\left(h_{t}^{*}({x}),  f_{T}\right)+\epsilon_{T}\left(h(x),  f_{T}\right) \\
			& \leq \underbrace{\epsilon_{\widetilde{T}}\left(h(\tilde{x}),  h_{t}^{*}({x})\right)}_{attack~defense}+\underbrace{\epsilon_{T}\left(h(x),  h_{t}^{*}({x})\right)}_{benign~maintenance}\\
			&+\underbrace{2 \epsilon_{T}\left(h_{t}^{*}({x}),  f_{T}\right)}_{ideal~classifier}
			\label{fun:1}
		\end{split}
\end{equation}
where $h_{t}^{*}$ denotes the ideal target classifier defined in Eq. (\ref{eqtargetclassifier}),  which can be easily obtained by pre-training an arbitrary UDA model induced from the generalization bound,  i.e. inequality~(\ref{udatheory}). 
\begin{proof}
Utilizing the triangle inequality,  we can easily derive the upper bound for the expected error of the adversarial and clean samples in the target domain. 

The first error term of the adversarial examples $\epsilon_{\widetilde{T}}\left(h(\tilde{x}),  f_{T}\right)$ in Inequality (\ref{fun:1}) is elaborated as:
\begin{equation}
\small
		\begin{split}
			\epsilon_{\widetilde{T}}\left(h(\tilde{x}),  f_{T}\right) &=\epsilon_{\widetilde{T}}\left(h(\tilde{x}),  f_{T}\right)+\epsilon_{T}\left(h_{t}^{*}({x}),  f_{T}\right)-\epsilon_{T}\left(h_{t}^{*}({x}),  f_{T}\right) \\ & \leq\left|\epsilon_{\widetilde{T}}\left(h(\tilde{x}),  f_{T}\right)-\epsilon_{T}\left(h_{t}^{*}({x}),  f_{T}\right)\right|+\epsilon_{T}\left(h_{t}^{*}({x}),  f_{T}\right) \\ & \leq \mathrm{E}_{\mathcal{D}_{\widetilde{T}}}\left[\left|h(\tilde{x})-f_{T}({x})-h_{t}^{*}({x})+f_{T}({x})\right|\right] \\
			&~~~+\epsilon_{T}\left(h_{t}^{*}({x}),  f_{T}\right) \\ &=\mathrm{E}_{\mathcal{D}_{\widetilde{T}}}\left[\left|h(\tilde{x})-h_{t}^{*}({x})\right|\right]+\epsilon_{T}\left(h_{t}^{*}({x}),  f_{T}\right) \\ &=\epsilon_{\widetilde{T}}\left(h(\tilde{x}),  h_{t}^{*}({x})\right)+\epsilon_{T}\left(h_{t}^{*}({x}),  f_{T}\right)
		\end{split}
	\label{bb}
\end{equation}

The second error term of the clean examples $\epsilon_{T}\left(h(x),  f_{T}\right)$ in Inequality (\ref{fun:1}) is derived as follows.
\begin{equation}
\small
		\begin{split}
			\epsilon_{T}\left(h(x),  f_{T}\right) &=\epsilon_{T}\left(h(x),  f_{T}\right)+\epsilon_{T}\left(h_{t}^{*}({x}),  f_{T}\right)-\epsilon_{T}\left(h_{t}^{*}({x}),  f_{T}\right) \\ & \leq\left|\epsilon_{T}\left(h(x),  f_{T}\right)-\epsilon_{T}\left(h_{t}^{*}({x}),  f_{T}\right)\right|+\epsilon_{T}\left(h_{t}^{*}({x}),  f_{T}\right) \\ & \leq \mathrm{E}_{\mathcal{D}_{T}}\left[\left|h({x})-f_{T}({x})-h_{t}^{*}({x})+f_{T}({x})\right|\right] \\
			&~~~+\epsilon_{T}\left(h_{t}^{*}({x}),  f_{T}\right) \\ &=\mathrm{E}_{\mathcal{D}_{T}}\left[\left|h({x})-h_{t}^{*}({x})\right|\right]+\epsilon_{T}\left(h_{t}^{*}({x}),  f_{T}\right) \\ &=\epsilon_{T}\left(h(x),  h_{t}^{*}({x})\right)+\epsilon_{T}\left(h_{t}^{*}({x}),  f_{T}\right)
		\end{split}
	\label{cc}
\end{equation}

By combining Inequality (\ref{bb}) and Inequality (\ref{cc}),  we can obtain the new generalization bound for URDA against adversarial perturbations,  i.e.,  Inequality (\ref{fun:1}).

The proof of the redefined upper bound for the expected target risk $\xi_{T,  \widetilde{T}}({h})$ in Theorem \ref{thm2} is done.
\end{proof}

\begin{remark}
	\textbf{Theorem~\ref{thm2} clarifies the theoretical upper bound of the URDA paradigm,  including three explanatory terms in Inequality (\ref{fun:1}).} \textbf{1)} The first term of the upper bound is the classification consistency error between the adversarial target samples on $h$ (i.e.,  $h(\widetilde{x})$) and the clean target samples on $h_{t}^{*}$ (i.e.,  $h_t^*(x)$),  which is termed as \textit{attack defense}. Existing work reveals that imposing a consistency constraint on adversarial and clean sample pairs can smooth the model and increase the adversarial robustness. \textbf{2)} The second term represents the classification consistency error for the clean target samples between $h(x)$ and $h_{t}^*(x)$,  which indicates that excellent classification performance on clean samples is maintained (termed as \textit{benign maintenance}) and the merit of the traditional UDA transfer training is inherited. In other words,  if we have trained a sufficiently reliable target domain classifier $h_{t}^{*}$ that is independent of $h$ by the standard UDA method,  then we only need to inherit its excellent classification performance,  rather than retraining the UDA models. These two error terms (i.e.,  model consistency errors) are indispensable conditional constraints to increase the adversarial robustness of the model. \textbf{3)} The third error term is naturally supposed to be small enough on clean data for an arbitrary non-robust UDA model (termed as ideal classifier). \textit{This implies a fundamental premise of our URDA paradigm and theory that the vanilla UDA models can well adapt to clean target data.}
\end{remark}

\begin{figure}[t]
	\centering
	\includegraphics[width=0.5\textwidth]{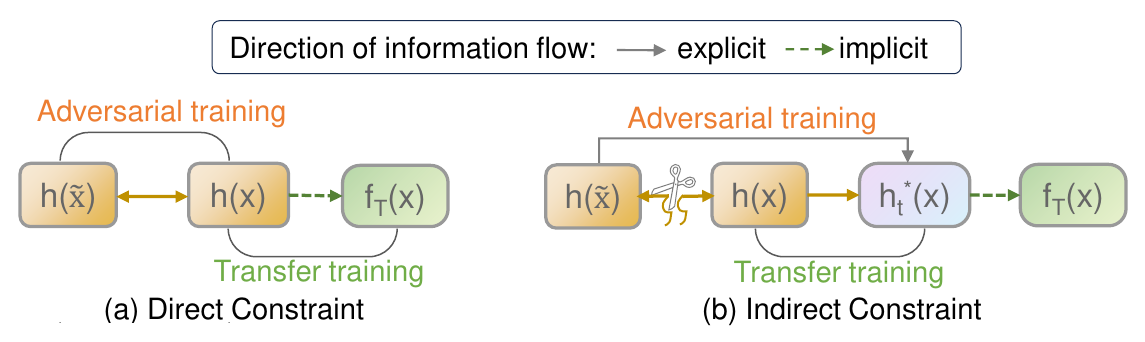}
	\caption{The information flow direction varies when imposing constraints between clean and adversarial samples. (a) Direct constraints on $h(\tilde{x})$ and ${h(x)}$,  which may damage transfer training and enforce the model to misclassify clean samples. (b) Indirect constraint on $h(\tilde{x})$ and ${h(x)}$ by introducing $h_t^*(x)$,  which ensures independent execution of transfer and adversarial training. This approach enhances robustness through adversarial training while preserving exceptional classification performance on clean samples through transfer training.}
\label{fig:flow}
\end{figure}

\begin{remark}
\textbf{Entanglement means direct constraint while disentanglement equals to indirect constraint}. With an ideal target domain classifier $h_{t}^{*}$ as the foundation,  the dilemma of entanglement between adversarial and transfer training is broken. Fig.~\ref{fig:flow} illustrates the information flow direction when imposing consistency constraints between clean and adversarial samples,  both directly and indirectly. In Fig.~\ref{fig:flow} (a),  direct constraints between $h(\tilde{x})$ and ${h(x)}$ force ${h(x)}$ to be influenced by both adversarial and transfer training. This jeopardizes transfer training and compromises clean accuracy. However,  introducing the ideal target classifier $h_{t}^{*}$ in Inequality (\ref{fun:1}) transforms the information flow from Fig.~\ref{fig:flow} (a) into (b). Then the entanglement between $h(\tilde{x})$ and ${h(x)}$ unravels,  rendering benign samples entirely independent of adversarial samples and mitigating potential negative transfer and model collapse. In Fig.~\ref{fig:flow} (b),  the supervision of $h_{t}^{*}$ for ${h(x)}$ preserves the classification performance of transfer training on benign samples,  while the supervision of $h_{t}^{*}$ for $h(\tilde{x})$ enhances the model's robustness to adversarial samples. Thus,  the ideal target classifier $h_{t}^{*}$ from a standard UDA model essentially functions as a teacher model,  distilling a robust student model with inherited teacher capabilities and endowed adversarial robustness. Therefore,  the key of URDA paradigm is to release the constraint from entanglement to disentanglement,  such that robustness is theoretically guaranteed.
\end{remark}

\begin{remark}
\textbf{The ideal target classifier $h_t^*$}. In Inequality (\ref{fun:1}), we introduce the ideal target classifier $h_t^*$ as a theoretical guarantee in order to disentangle the transfer training (i.e., DA term) and adversarial training (i.e., robustness term) in the URDA bound. It is important to emphasize that $h_t^*$ is not required to be perfectly achieved in practice. Instead, in our implementation, $h_t^*$ is approximated by a vanilla UDA model (non-robust teacher), which provides sufficiently accurate hypotheses on clean samples of the target domain. Although this teacher model is not robust, it serves to approximate the target risk minimizer and acts as a practical proxy for $h_t^*$ to guide the robust student model learning. $h_t^*$ and the robust student model $h$ share the same backbone and parameterization, and therefore belong to the same hypothesis space $\mathcal{H}$. The basic philosophy is: the teacher model acquires domain adaptation capability via a standard UDA, while the student model inherits this ability and further acquires robustness via distillation. Therefore, their error bounds can be analyzed under a unified hypothesis.
\end{remark}

\subsection{The Proposed URDA Algorithm: DART}
Theorem \ref{thm2} establishes the generalization bound of URDA for the first time,  which focuses on the real-time minimization of $\epsilon_{\widetilde{T}}\left(h(\tilde{x}),  h_{t}^{*}(x)\right)$ and $\epsilon_{T}\left(h(x),  h_{t}^{*}(x)\right)$ as delineated in Inequality (\ref{fun:1}). Based on the URDA theoretical upper bound, we formularize a URDA algorithm, i.e. DART. As depicted in Fig.~\ref{fig:2}, the first step is to obtain a reliable target classifier adaptable to clean target samples and the second step is to robustify the model obtained in the first step. Concretely,  DART consists of two disentangled training steps elaborated as follows:

\textbf{Step 1: Classical UDA Pre-training}. Pre-train a vanilla UDA model off-line to acquire a classification model ${M}_{uda}$ according to the target error upper bound in Inequality (\ref{udatheory}),  which then determines the target classifier $h_{t}^{*}$ in the proposed Inequality (\ref{fun:1}).

\textbf{Step 2: Robustification Post-training}. By distilling the UDA classification model ${M}_{uda}$,  a robust model ${M}_{rob}$ is learned. To reduce the discrepancies between the two models,  we impose consistency constraints at the feature and prediction levels. MSE (Mean Square Error) loss is used to constrain the feature consistency,  while Kullback-Leibler (KL) divergence is used to constrain the prediction consistency. This aims to minimize the upper bound of the proposed Inequality (\ref{fun:1}).

By connecting the above two steps with the new upper bound in Inequality (\ref{fun:1}),  we can obtain
\begin{equation}
\small
		\begin{split}
			\xi_{T, \widetilde{T}}\left(h,  f_{T}\right)
			& \leq \underbrace{\epsilon_{\widetilde{T}}\left(h(\tilde{x}),  h_{t}^{*}\right)+\epsilon_{T}\left(h(x),  h_{t}^{*}\right)}_{\emph{Step 2}}+ \underbrace{2 \epsilon_{T}\left(h_{t}^{*},  f_{T}\right)}_{\emph{Step 1}}
		\end{split}
	\label{dd}
\end{equation}
The specific technical details of Step 1 and Step 2 for pre- and post- training are elaborated as follows.
\subsubsection{Classical UDA Pre-training (Step 1)}
As shown in Fig.~\ref{fig:2},  we first employ an arbitrary UDA method to train and obtain a non-robust classification model ${M}_{ {uda}}$ by minimizing the third term $\epsilon_{T}\left(h_{t}^{*},  f_{T}\right)$. This is theoretically guaranteed by the Inequality (\ref{udatheory}). In practice,  we can attain the non-robust classification model ${M}_{ {uda}}$ by any of the UDA methods. The loss function $\mathcal{L}_{uda}$ be expressed by the following formula.
\begin{equation}
\mathcal{L}_{uda}=\mathcal{L}_{cls} + \mathcal{L}_{da}
	\label{8}
\end{equation}
Specifically,  $\mathcal{L}_{cls}$ is a classification loss,  which uses the standard cross-entropy loss $\mathcal{L}_{ce}$ and is formulated as:
\begin{equation}
\begin{split}
\mathcal{L}_{cls} & =  \mathbb{E}_{x_i^{s}\sim\mathcal{D}_S}\mathcal{L}_{ce}\left({M}_{uda}\left({x}_{i}^s\right),  y_{i}\right)\  
\\
& = \frac{1}{m} \sum_{i=1}^n \mathcal{L}_{ce}\left({C}_{uda}(E_{uda}\left({x}_{i}^s\right)),  y_{i}\right)
\end{split}
	\label{l_cls}
\end{equation}
where $C_{uda}(\cdot)$ represents the classifier and $E_{uda}(\cdot)$ means the feature encoder of the model ${M}_{ {uda}}$ as shown in Fig.~\ref{fig:2}. $\mathcal{L}_{da}$ is the classic adversarial domain alignment loss,  which can be formulated as:
\begin{equation}
\begin{split}
\mathcal{L}_{da}  = ~& \mathbb{E}_{x_i^{s}\sim\mathcal{D}_S}log[D_{uda}(E_{uda}(x_i^s)) +\\
&\mathbb{E}_{x_j^{t}\sim\mathcal{D}_T}log [1-[D_{uda}(E_{uda}(x_j^t))]\\
=~&\frac{1}{m} \sum_{i=1}^m log[D_{uda}(E_{uda}(x_i))]+
\\
& \frac{1}{n} \sum_{j=1}^n log [1-[D_{uda}(E_{uda}(x_j))]
\end{split}
	\label{l_cls}
\end{equation}
where $m$ and $n$ refer to the number of the source domain and target domain in training set,  respectively. $D_{uda}(\cdot)$ means the domain discriminator. For UDA pre-training,  we adapt common domain adversarial learning to update the model. The final objective function is 
\begin{equation}\max_{\theta_{E_{uda}}, \theta_{C_{uda}}}~~
\min_{\theta_{D_{uda}}} ~~\mathcal{L}_{uda}
\label{9}
\end{equation}
where the $\theta_{E_{uda}}$,  $\theta_{C_{uda}}$ and $\theta_{D_{uda}}$ denote the parameters of feature extractor,  UDA classifier and domain discriminator,  respectively. Note that our URDA paradigm is model agnostic,  so existing or customized UDA model can be considered in designing URDA algorithm. In this paper,  we focus on the construction of paradigm,  theory and algorithm for URDA,  therefore,  in experiments,  we adopt the existing UDA models (e.g.,  CDAN,  MCC and MDD) in Step 1 to validate the superiority of the URDA baseline algorithm,  i.e.,  DART.



\subsubsection{Robustification Post-training (Step 2)}
The second step is to minimize two loss functions $\mathcal{L}_{\widetilde{T}}$ and $\mathcal{L}_{T}$ by performing model distillation. This is coined as robustification post-training by frozen the pre-trained UDA model in Step 1. $\mathcal{L}_{\widetilde{T}}$ is used to ensure model's robustness towards adversarial samples and $\mathcal{L}_{T}$ is used to ensure classification performance on clean samples. The objective function of Step 2 to characterize the upper bound of Inequality (\ref{dd}) is formulated as:
\begin{equation}
		\begin{aligned}
			&{\arg \underset{\theta_{\text{rob}}}\min } ~\mathcal{L}_{\widetilde{T}}\left(x^{t},  \tilde{x}^{t}\right)+\mathcal{L}_{T}\left(x^{t}\right), \\
			&where~~~\tilde{{x}}^{t}=\underset{\left\|\tilde{x}^{t}-x^{t}\right\|_{p}\leq \epsilon}{\arg \max } \mathcal{L}_{c e}(g(x^{t}),  \hat{y})\\
		\end{aligned}
\label{equationstep2}
\end{equation}
where $\mathcal{L}_{ce}$ means cross-entropy loss and $\epsilon>0$ is a small constant.
We utilize I-FGSM \cite{kurakin2018adversarial},  a non-targeted attack method,  to obtain adversarial samples. Note that other adversarial attackers can also be used,  such as PGD \cite{madry2017towards} and AutoAttack \cite{croce2020reliable},  which have been tested in experiments. Due to lacking ground-truth target labels,  we use the pseudo target label $\hat{y}$ (predicted labels of clean target samples by the pretrained UDA classifier) to generate adversarial samples. Note that \textit{the pseudo target labels impose little impact because they resort to generating fake samples of wrong classes}. We also confirm that in experimental discussion. Specifically,  we inherit the pre-trained UDA model and conduct on-the-fly robustification post-training to minimize the first two terms in Inequality (\ref{dd}).

\textbf{Minimization of $\epsilon_{\widetilde{T}}\left(h(\tilde{x}),  h_{t}^{*}(x)\right)$ in Inequality (\ref{dd})}. The loss $\mathcal{L}_{\widetilde{T}}$ in Problem (\ref{equationstep2}) characterizes this error,  which makes use of the KL loss to ensure that the prediction $\widetilde{P}_{i}^{t}$ of attack samples $\tilde{{x}}_{i}^{t}$ on robust student model ${M}_{ {rob}}$ and the prediction ${P}_{i}^{t}$ of benign samples ${{x}}_{i}^{t}$ on non-robust teacher model ${M}_{ {uda}}$ are as consistent as possible.
To enhance the precision of the supervisory signal from the teacher network,  we incorporate predictive entropy,  denoted as $H({x}_{i}^{t})=-\frac{1}{\log{c}}\sum_{j=1}^{c}P_{ij}\log P_{ij}$,  where $c$ represents the number of classes and $P_i$ is the probability distribution for predicting a sample $x_i^t$ (i.e.,  $P_i=[P_{i1}, P_{i2}, ..., P_{ic}]$). The distillation-based loss $\mathcal{L}_{\widetilde{T}}$ can be formulated as follows:
\begin{equation}
\small
		\begin{split}
			\mathcal{L}_{\widetilde{T}}=\sum_{i=1}^{n} (1 \!\!-\!\! H({x}_{i}^{t}))K L\left(C_{uda}\left(E_{{uda}}\left({x}_{i}^{t}\right),  C_{rob}\left(E_{{rob}}\left(\tilde{{x}}_{i}^{t}\right)\right)\right) \right)
		\end{split}
	\label{ee}
\end{equation}
where $C(\cdot)$ represents the classifier and $E(\cdot)$ means the feature encoder as shown in Fig.~\ref{fig:2}. The loss function (\ref{ee}) is to minimize $\epsilon_{\widetilde{T}}\left(h(\tilde{x}),  h_{t}^{*}\right)$ and ensure the robustness of ${M}_{ {rob}}$ to adversarial target samples (i.e.,  attack defense).

\textbf{Minimization of $\epsilon_{T}\!\left(h(x),  h_{t}^{*}(x)\right)$ in Inequality (\ref{dd})}. The loss $\mathcal{L}_{T}$ in Problem (\ref{equationstep2}) characterizes this error. To better inherit the performance of ${M}_{ {uda}}$ and ensure the classification ability of ${M}_{ {rob}}$ on clean samples,  the loss function $\mathcal{L}_{T}$ constrains the features and prediction consistency between the two models simultaneously. Specifically,  $\mathcal{L}_{T}$ uses MSE loss and KL loss to constrain the clean sample ${{x}}_{i}^{t}$ between the non-robust UDA model ${M}_{ {uda}}$ and the robust model ${M}_{ {rob}}$ learned on the fly,  which can be expressed as:
\begin{equation}
\small
		\begin{split}
			\mathcal{L}_{T}&=MSE\left(F_{uda},  F_{rob}\right)+KL\left(P_{uda},  P_{rob}\right)\\
			&=\sum_{i=1}^{n} [MSE\left(E_{u d a}\left({x}_{i}^{t}\right),  E_{r o b}\left({x}_{i}^{t}\right)\right)\\
			&+(1-H({x}_{i}^{t}))KL\left(C_{u d a}\left(E_{u d a}\left({x}_{i}^{t}\right)\right),  C_{r o b}\left(E_{r o b}\left({x}_{i}^{t}\right)\right)\right)]
		\end{split}
	\label{distll_clean}
\end{equation}
where $F$ means the feature representations from encoder $E$ and $P$ means the the prediction logits from classifier $C$.
These consistency constraints minimize $\epsilon_{T}\left(h(x),  h_{t}^{*}(x)\right)$ and ensure classification accuracy on clean (benign) target samples (i.e.,  benign maintenance).
\subsubsection{Overall Training Objective and Algorithm}
The overall optimization objective of our DART algorithm for URDA consists of two steps: UDA pre-training step and Robustification post-training step,  which are disentangled and independent. Specifically,  DART is formulated as:
\begin{equation}
		\mathcal{L}_{DART}=
		\begin{cases}
			&{\arg \underset{\theta_{\text {uda }}}\min }~\mathcal{L}_{uda}, \\
			&{\arg \underset{\theta_{\text {rob }}}\min }~ \mathcal{L}_{\widetilde{T}}\left(x^{t},  \tilde{x}^{t}\right)+\mathcal{L}_{T}\left(x^{t}\right)\\
		\end{cases}
	\label{13}
\end{equation}

In short,  the DART paradigm disentangles the transfer training phase (Step 1) from the robust training phase (Step 2),  and essentially overcomes the interactive entanglement problem. DART is then guaranteed to be effective and robust for both clean and adversarial samples by following the proved theoretical generalization bound in Theorem \ref{thm2}. 

The training procedure of DART is summarised in Algorithm \ref{algorithm},  which is embarrassingly easy to implement yet with theoretical guarantees.


\begin{algorithm}[t]
	\caption{:~DART (A theoretically guaranteed training algorithm of URDA) }
	\label{algorithm}
	\begin{algorithmic}[1]
		\REQUIRE Labeled source domain $\mathcal{D}_S=\left\{\left(x_{i}^{s},  y_{i}^{s}\right)\right\}_{i=1}^{m}$;  unlabeled target domain $\mathcal{D}_T=\left\{x_{i}^{t}\right\}_{i=1}^{n}$; a vanilla UDA model (e.g.,  \textbf{CDAN}~\cite{long2018conditional},  \textbf{MCC}~\cite{DBLP:journals/corr/abs-1912-03699} and \textbf{MDD}~\cite{zhang2019bridging},  etc.); batch size $B$; learning rate $lr$;  iteration number $N$.
		\ENSURE The URDA model $M_{rob}$.\\
		\textbf{Step-1 (UDA):}
		Pretrain a UDA model $M_{uda}$ through Eqs. (\ref{8})-(\ref{9}). \textcolor{mygreen}{\textit{\# get the ideal target classifier $h^{*}_t$}}\\
		\textbf{Step-2 (Robustification):}
		Repeat until converge or reach maximum iterations:\\
		\lowercase\expandafter{\romannumeral 1}. Generate pseudo labels $\hat{y}$ for target data $x_{t}$\ by $M_{uda}$;  \textcolor{mygreen}{\textit{\# pseudo target label prediction}} \\
		\lowercase\expandafter{\romannumeral 2}. Generate adversarial examples of target data $\tilde{x}^{t}$ based on the pseudo labels through I-FGSM~\cite{kurakin2018adversarial} attacker;  \textcolor{mygreen}{\textit{\# other types of attackers can also be used in implementation}}\\
		\lowercase\expandafter{\romannumeral 3}. Update the parameters of $M_{rob}$ with Eq. (\ref{13}).  \textcolor{mygreen}{\textit{\# perform robustification post-training for URDA}}\\
       \textbf{End}
	\end{algorithmic}
\end{algorithm}

\begin{table*}[ht]
	\centering
\caption{Accuracy (clean,  \%) and adversarial robustness (Adv.,  \%) on Office-31. The pink-highlighted cells denote the upper bounds of clean accuracy. Notably, the dashed lines for RFA and RAT mean the results are absent in their original paper.}
\vspace{-5mm}
\label{table:Office-31}
\begin{center}
\setlength{\tabcolsep}{8pt}
\renewcommand{\arraystretch}{0.9}
	\resizebox{\linewidth}{!}{
\begin{tabular}{p{0.1mm} c c c c c c c c c c c c|c c  }
			\hline
			\multicolumn{1}{ l|}{\multirow{2}*{Method}}
			& \multicolumn{2}{c}{A$\rightarrow$D}
			& \multicolumn{2}{c}{A$\rightarrow$W}
			& \multicolumn{2}{c}{D$\rightarrow$A}
			& \multicolumn{2}{c}{D$\rightarrow$W}
			& \multicolumn{2}{c}{W$\rightarrow$A}
			& \multicolumn{2}{c|}{W$\rightarrow$D}
			& \multicolumn{2}{c}{Avg.}       \\
			
			\multicolumn{1}{ l|}{}
			& clean    & Adv.     & clean    & Adv.
			& clean    & Adv.     & clean    & Adv.
			& clean    & Adv.     & clean    & Adv.
			& clean    & Adv.     \\
			\hline
			
			\multicolumn{1}{ l|}{RFA~\cite{awais2021adversarial}}
			& -                 & -
			& -                 & -
			& -                 & -
			& -                 & -
			& -                 & -
			& -                 & -
			& 84.2              & 74.3     \\
            \multicolumn{1}{ l|}{\textcolor{black}{RAT}~\cite{xiao2023adversarially}}
			& -                 & -
			& -                 & -
			& -                 & -
			& -                 & -
			& -                 & -
			& -                 & -
			& \textcolor{black}{87.7}              & \textcolor{black}{87.7}   \\
			\multicolumn{1}{ l|}{UCAT~\cite{10043655}}
			& {95.2}     & 67.9
			& 89.2              & 68.4
			& 78.0              & 48.6
			& 97.1              & 70.5
			& {78.0}     & 49.7
			& 98.7              & 70.6
			& {89.4}     & 62.6     \\
			\multicolumn{1}{ l|}{SRoUDA~\cite{zhu2022srouda}}
			& 91.9              & 85.5
			& {95.9}     & 84.7
			& 72.5              & 57.2
			& 96.7              & 83.9
			& 67.1              & 57.4
			& 100.0             & 88.7
			& 87.3              & 75.8     \\
\hline

	\end{tabular}}
\end{center}
\end{table*}

\begin{table*}[ht]
	\caption{Accuracy (clean,  \%) and adversarial robustness (Adv.,  \%) on Office-Home and Amazon Reviews datasets. The pink-highlighted cells denote the upper bounds of clean accuracy. Notably, the dashed lines mean no result in the original paper.}
 \vspace{-5mm}
\label{table:Office-home}
\begin{center}
\setlength{\tabcolsep}{0.2pt}
\renewcommand{\arraystretch}{1.2}
	\resizebox{\linewidth}{!}{
		\begin{tabular}{p{0.1mm} c c c c c c c c c c c c c c c c c c c c c c c c|c c}
			\hline
			\multicolumn{1}{ l|}{\multirow{2}*{Office-Home}}
			& \multicolumn{2}{c}{Ar $\rightarrow$ Cl}
			& \multicolumn{2}{c}{Ar $\rightarrow$ Pr}
			& \multicolumn{2}{c}{Ar $\rightarrow$ Rw}
			& \multicolumn{2}{c}{Cl $\rightarrow$ Ar}
			& \multicolumn{2}{c}{Cl $\rightarrow$ Pr}
			& \multicolumn{2}{c}{Cl $\rightarrow$ Rw}
			& \multicolumn{2}{c}{Pr $\rightarrow$ Ar}
			& \multicolumn{2}{c}{Pr $\rightarrow$ Cl}
			& \multicolumn{2}{c}{Pr $\rightarrow$ Rw}
			& \multicolumn{2}{c}{Rw $\rightarrow$ Ar}
			& \multicolumn{2}{c}{Rw $\rightarrow$ Cl}
			& \multicolumn{2}{c|}{Rw $\rightarrow$ Pr}
			& \multicolumn{2}{c}{Avg} \\
			
			\multicolumn{1}{ l|}{}
			& clean    & Adv.     & clean    & Adv.
			& clean    & Adv.     & clean    & Adv.
			& clean    & Adv.     & clean    & Adv.
			& clean    & Adv.     & clean    & Adv.
			& clean    & Adv.     & clean    & Adv.
			& clean    & Adv.     & clean    & Adv.
			& clean    & Adv.   \\
			\hline
			
			\multicolumn{1}{ l|}{RFA~\cite{awais2021adversarial}}
			& -        & -        & -        & -
			& -        & -        & -        & -
			& -        & -        & -        & -
			& -        & -        & -        & -
			& -        & -        & -        & -
			& -        & -        & -        & -
			& 65.4     & 51.1 \\
            \multicolumn{1}{ l|}{\textcolor{black}{RAT}~\cite{xiao2023adversarially}}
			& -        & -        & -        & -
			& -        & -        & -        & -
			& -        & -        & -        & -
			& -        & -        & -        & -
			& -        & -        & -        & -
			& -        & -        & -        & -
			& \textcolor{black}{69.7}     & \textcolor{black}{69.3} \\
\hline
	\end{tabular}}
\end{center}
\end{table*}

\begin{table*}[ht]
\caption{Accuracy (clean,  \%) and adversarial robustness (Adv.,  \%) on VisDA-2017. The pink-highlighted cells denote the upper bounds of clean accuracy. Notably, the dashed lines for RFA mean no result in their original paper.}
\vspace{-5mm}
\label{table:VisDA-2017}
\begin{center}
\setlength{\tabcolsep}{0.2pt}
\renewcommand{\arraystretch}{1.2}
\resizebox{\linewidth}{!}{
	\begin{tabular}{p{0.4mm} c c c c c c c c c c c c c c c c c c c c c c c c|c c}
		\hline
		\multicolumn{1}{ l|}{\multirow{2}*{Method}}
		& \multicolumn{2}{c}{plane}
		& \multicolumn{2}{c}{bcycl}
		& \multicolumn{2}{c}{bus}
		& \multicolumn{2}{c}{car}
		& \multicolumn{2}{c}{horse}
		& \multicolumn{2}{c}{knife}
		& \multicolumn{2}{c}{mcycl}
		& \multicolumn{2}{c}{person}
		& \multicolumn{2}{c}{plant}
		& \multicolumn{2}{c}{sktbrd}
		& \multicolumn{2}{c}{train}
		& \multicolumn{2}{c|}{truck}
		& \multicolumn{2}{c}{Avg}  \\
		
		\multicolumn{1}{ l|}{}
		& clean    & Adv.     & clean    & Adv.
		& clean    & Adv.     & clean    & Adv.
		& clean    & Adv.     & clean    & Adv.
		& clean    & Adv.     & clean    & Adv.
		& clean    & Adv.     & clean    & Adv.
		& clean    & Adv.     & clean    & Adv.
		& clean    & Adv. \\
		\hline

        \multicolumn{1}{ l|}{RFA~\cite{awais2021adversarial}}
		& -        & -        & -        & -
		& -        & -        & -        & -
		& -        & -        & -        & -
		& -        & -        & -        & -
		& -        & -        & -        & -
		& -        & -        & -        & -
		& 72.9     & 47.7 \\
\hline
\end{tabular}}
\end{center}
\end{table*}

\begin{table*}[t]
	\caption{Accuracy (clean,  \%) and adversarial robustness (Adv.,  \%) on DomainNet. The pink-highlighted cells denote the upper bounds of clean accuracy. Notably, the dashed lines mean no result in the original paper.}
 \vspace{-3mm}
\label{table:DomainNet}
\begin{center}
\setlength{\tabcolsep}{1.5pt}
\renewcommand{\arraystretch}{1}
	\resizebox{\linewidth}{!}{
		\begin{tabular}{p{0.5mm} c c c c c c c c c c c c ||l |c c c c c c c c c c c c }
			\hline

\multicolumn{1}{ c|}{\textcolor{black}{Source}}
			& \multicolumn{2}{c}{clp}
			& \multicolumn{2}{c}{inf}
			& \multicolumn{2}{c}{pnt}
			& \multicolumn{2}{c}{qdr}
			& \multicolumn{2}{c}{rel}
			& \multicolumn{2}{c||}{skt}
			& \multicolumn{1}{c|}{}
			& \multicolumn{2}{c}{clp}
			& \multicolumn{2}{c}{inf}
			& \multicolumn{2}{c}{pnt}
			& \multicolumn{2}{c}{qdr}
			& \multicolumn{2}{c}{rel}
			& \multicolumn{2}{c}{skt}     \\
			
			\multicolumn{1}{ c|}{\textcolor{black}{-only}}
			& clean    & Adv.     & clean    & Adv.
			& clean    & Adv.     & clean    & Adv.
			& clean    & Adv.     & clean    & Adv.
			& \multicolumn{1}{ c|}{\textcolor{black}{MIRoUDA}}
			& clean    & Adv.     & clean    & Adv.
			& clean    & Adv.     & clean    & Adv.
			& clean    & Adv.     & clean    & Adv.     \\
			\hline

			\multicolumn{1}{c|}{clp}
			& -        & -
			& -    & -
			& \textcolor{black}{25.4 }   & \textcolor{black}{10.3}
			& -     & -
			& \textcolor{black}{36.4}     & \textcolor{black}{16.8}
			& \textcolor{black}{30.2 }   & \textcolor{black}{21.8 }  
            & \multicolumn{1}{ c|}{clp}
			& -        & -        & -     & -
			& \textcolor{black}{40.3}     & \textcolor{black}{20.8}     & -     & -
			& \textcolor{black}{53.2 }    & \textcolor{black}{31.6}     & \textcolor{black}{42.4}     & \textcolor{black}{32.7}
            \\
			
			\multicolumn{1}{ c|}{inf}
			&-     & -     & -        & -
			& -     & -     & -      & -
			& -     & -     & -     & -
			& \multicolumn{1}{c|}{inf}
			& -     & -
			& -        & -
			& -     &-
			& -    & -
			& -     & -
			& -    & -     \\
\hline
	\end{tabular}}
\end{center}
\end{table*}

\section{Experiments}\label{sec4}
\subsection{Datasets}
\textbf{Office-31}~\cite{saenko2010adapting} is a widely used domain adaptation benchmark,  which contains 4, 110 images of 31 categories shared by three distinct domains: \textit{Amazon} (A),  \textit{Webcam} (W) and \textit{Dslr} (D).

\textbf{Office-Home}~\cite{venkateswara2017deep} consists of 15, 500 images in 65 object classes,  from four different domains: \textit{Artistic images} (Ar),  \textit{Clip Art} (Cl),  \textit{Product images} (Pr) and \textit{Real-World images} (Rw).

\textbf{VisDA-2017}~\cite{peng2017visda} is a large-scale dataset with two distinct domains: \textit{Synthetic} and \textit{Real},  sharing 12 classes.

\textbf{DomainNet}~\cite{Peng_2019_ICCV} contains about 0.6 million images of 345 classes in six domains: \textit{Clipart} (clp),  \textit{Infograph} (inf),  \textit{Painting} (pnt),  \textit{Quickdraw} (qdr),  \textit{Real} (rel),  \textit{Sketch} (skt).

\textbf{Amazon Reviews}~\cite{blitzer2007biographies} contains four domains: Book (B), DVD (D), Electronics (E) and Kitchen (K). For each domain, there are 1000 positive and 1000 negative labeled reviews, and 4000 randomly chosen unlabeled data.

\subsection{Experimental Setup}
For Office-31 and Office-Home datasets,  6 and 12 cross-domain tasks are tested,  respectively. For VisDA-2017,  we take \textit{Synthetic} as source and \textit{Real} as target,  i.e.,  \textit{Synthetic}$\rightarrow$\textit{Real}. For DomainNet,  we take six domains i.e.,  \textit{Clipart} (clp),  \textit{Infograph} (inf),  \textit{Painting} (pnt),  \textit{Quickdraw} (qdr),  \textit{Real} (rel) and \textit{Sketch} (skt)) with 345 classes to conduct experiments in 30 cross-domain tasks. In each adaptation task,  the images and their class labels in source domain are utilized,  while for target domain,  only images are used.

\textbf{Compared methods.} To verify the effectiveness of the proposed DART algorithm,  we choose five UDA baselines, including three classic methods: \textbf{CDAN}~\cite{long2018conditional},  \textbf{MCC}~\cite{DBLP:journals/corr/abs-1912-03699}, \textbf{MDD}~\cite{zhang2019bridging} and two recent pseudo-labeled UDA methods: \textbf{UPA}~\cite{chen2024uncertainty}, \textbf{PL-Mix}~\cite{kong2024unsupervised} to conduct robustification experiments. We compare four methods in the experiments: None,  +AT,  +Trades and +DART. ``None'' represents the baseline UDA model without any robustification training. Both ``+AT'' and ``+Trades'' denote adversarial training by adding adversarial samples,  which follow the UDA+VAT paradigm,  while ``+DART'' follows the proposed URDA paradigm. For details of ``+AT'',  on source domain,  the label information is used to supervise clean source samples and corresponding adversarial source samples,  respectively. Besides,  clean target samples are aligned with clean source samples and the adversarial target samples are aligned with adversarial source samples. The overall training objectives include four parts,  i.e.,  two domain transfer losses and two classification losses. Note that the transfer loss adopts classic domain adversarial learning (DAL) loss between feature generator and domain discriminator. The classification loss adopts standard cross-entropy loss. For details of ``+Trades'',  it performs adversarial training based on the idea of \cite{zhang2019theoretically},  where we utilize source labels to supervise clean source samples and corresponding adversarial source samples based on cross-entropy loss. It also minimizes the predictive difference between clean target samples and adversarial target samples by KL loss. ``+DART''  represents the proposed URDA training algorithm. Besides, given the recent focus of UDA community on attack robustness, we compare the corresponding works, including ARTUDA~\cite{Lo_2022_ACCV}, RFA~\cite{awais2021adversarial}, RAT~\cite{xiao2023adversarially}, UCAT~\cite{10043655}, SRoUDA~\cite{zhu2022srouda}, MIRoUDA~\cite{yin2024towards} and Dart~\cite{wang2025dart}. Furthermore,  we explore three powerful attack methods including I-FGSM \cite{kurakin2018adversarial},  PGD \cite{madry2017towards} and AutoAttack \cite{croce2020reliable} in order to show the generalization ability to different attack methods.

\textbf{Implementation Details.} We implement our method based on PyTorch. In \emph{Step 1 (off-line)},  we adhere to the same experimental setup as the baseline. For Office-31 and Office-Home,  we utilize ResNet50~\cite{he2016deep} as the backbone,  with a batch size of 36. For VisDA-2017 and DomainNet,  ResNet101~\cite{he2016deep} is employed as the backbone,  and the batch size is set to 18. In \emph{Step 2 (on-the-fly)},  the same backbones utilized in the pre-training stage are employed for robustness training. During model distillation,  we set the temperature parameter $\tau=2$ to calculate KL loss. Other settings are the same as the baseline. For the samples of training and testing,  the entire target samples are involved in robustness training and testing. For the generation way of adversarial samples,  adversarial samples generated by pseudo-labels are used for robustness training,  and adversarial samples generated with ground-truth labels are used for robustness testing. We extend the attack settings in traditional adversarial defense and increase the maximum attack limit to 0.1 for a more powerful attack. There are 40 attack perturbation steps in total and the size of each perturbation step is 0.05.

\subsection{Results}
The results on Office-31,  Office-Home, Amazon Reviews, VisDA-2017 and DomainNet with 4 baselines are presented in Tables \ref{table:Office-31},  \ref{table:Office-home},  \ref{table:VisDA-2017},  and \ref{table:DomainNet},  respectively. Table \ref{table:5}, \ref{table:6} and \ref{table:DomainNetaa} show the results on different attack methods. The pink-highlighted cells denote the upper bounds (i.e.,  None) on the performance of clean target samples. In general,  for different baselines (i.e.,  CDAN,  MCC,  and MDD),  non-robust model (i.e.,  None) achieves the best performance on clean target samples,  but the worst performance on the adversarial samples due to ignoring the safety of the model. +AT and +Trades proposed for the safety of the model improve the classification performance of the model on the adversarial samples,  i.e.,  adversarial robustness,  but greatly reduce the classification accuracy of the clean target samples. This is because they ignore the entanglement between transfer training and adversarial training. In contrast,  by suppressing the entanglement between adversarial training and transfer training,  the proposed DART achieves best performance on both the clean and adversarial samples,  i.e.,  $\text{None}\!>\!+\text{DART}\!>\!+\text{Trades}\!>\!+\text{AT}$ for clean samples and $+\text{DART}\!>\!+\text{Trades}\!>\!+\text{AT}\!>\!\text{None}$ for adversarial samples. This indicates two points: 1) the non-robust UDA achieves the best clean accuracy, and DART does not deteriorate the clean accuracy. 2) DART achieves the best robustness to adversarial samples.

\begin{table}[t]
	\centering
    \caption{UDA robustness comparison against different attacks on Office-31. Baseline is CDAN. The pink-highlighted cells denote the upper bounds of clean accuracy.}
	\label{table:5}
	\resizebox{\linewidth}{!}
    {
		\begin{tabular}{p{10mm} c| cccccc| c  }
	\hline
	\multicolumn{1}{ l}{Robustification method}
	& \multicolumn{1}{|l|}{Attack method}
	& {A$\rightarrow$D}
	& {A$\rightarrow$W}
	& {D$\rightarrow$A}
	& {D$\rightarrow$W}
	& {W$\rightarrow$A}
	& {W$\rightarrow$D}
	& {Avg.}       \\
    \hline

\multirow{2}{*}{\textcolor{black}{ARTUDA~\cite{Lo_2022_ACCV}}}
    &\multicolumn{1}{ |l|}{\textcolor{black}{None (clean)}}
    &\textcolor{black}{47.8} &\textcolor{black}{47.7} &\textcolor{black}{42.9} &\textcolor{black}{88.8} &\textcolor{black}{60.0} &\textcolor{black}{94.2} &\textcolor{black}{63.6} \\
	
    &\multicolumn{1}{|l|}{\textcolor{black}{AutoAttack}}
	  & \textcolor{black}{45.6}   & \textcolor{black}{45.2}
	& \textcolor{black}{33.1}     & \textcolor{black}{86.6}
	& \textcolor{black}{36.7}     & \textcolor{black}{91.6}
	& \textcolor{black}{56.5}                  \\        
	\hline

\multirow{2}{*}{\textcolor{black}{RFA~\cite{awais2021adversarial}}}
    &\multicolumn{1}{ |l|}{\textcolor{black}{None (clean)}}
    &\textcolor{black}{78.5} &\textcolor{black}{73.8} &\textcolor{black}{62.3} & \textcolor{black}{98.2} & \textcolor{black}{61.0} & \textcolor{black}{99.2} &\textcolor{black}{78.9} \\
	
    &\multicolumn{1}{|l|}{\textcolor{black}{AutoAttack}}
	  & \textcolor{black}{45.2}   & \textcolor{black}{33.1}
	& \textcolor{black}{46.6}     & \textcolor{black}{79.9}
	& \textcolor{black}{44.0 }    & \textcolor{black}{81.5}
	& \textcolor{black}{55.0}                  \\        
	\hline

\multirow{2}{*}{\textcolor{black}{SRoUDA~\cite{zhu2022srouda}}}
    &\multicolumn{1}{ |l|}{\textcolor{black}{None (clean)}}
    &\textcolor{black}{90.0} &\textcolor{black}{91.6} &\textcolor{black}{49.4} & \textcolor{black}{98.0} & \textbf{\textcolor{black}{71.9}} & \textcolor{black}{98.6} &\textcolor{black}{83.2} \\
	
    &\multicolumn{1}{|l|}{\textcolor{black}{AutoAttack}}
	  & \textcolor{black}{85.5}   & \textcolor{black}{90.6}
	& \textcolor{black}{22.4}   & \textcolor{black}{90.3}
	& \textcolor{black}{65.7}     & \textcolor{black}{\textbf{98.0}}
	& \textcolor{black}{75.4}                 \\        
	\hline

\multirow{4}{*}{CDAN (None)}
    &\multicolumn{1}{ |l|}{None (clean)}
    &\cellcolor{hfxpink}91.8 & \cellcolor{hfxpink}94.2&\cellcolor{hfxpink}73.4&\cellcolor{hfxpink}98.5&\cellcolor{hfxpink}71.4&\cellcolor{hfxpink} 99.8&\cellcolor{hfxpink}88.2\\
    &\multicolumn{1}{ |l|}{I-FSGM}
    
	& 31.1    & 24.2
	& 36.1     & 38.2
	& 42.8     & 35.9
	& 34.7                \\
	  &\multicolumn{1}{ |l|}{PGD}
	    & 30.7    & 26.7
	& 34.6     & 45.6
	& 41.9     & 34.9
	& 35.7                \\
	& \multicolumn{1}{ |l|}{AutoAttack}
	        & 11.7    & 10.3
	& 18.4     & 13.4
	& 11.1     & 9.8
	& 12.5                \\        
	\hline
	\multirow{4}{*}{CDAN + AT}
    &\multicolumn{1}{ |l|}{None (clean)}
    &80.5 &62.5 &35.8 &95.6 &45.4 &99.8 &69.9 \\
    &\multicolumn{1}{|l|}{I-FSGM}
    & 45.9    & 48.2
	& 41.3     & 80.6
	& 46.4     & 92.2
	& 59.1                  \\
	&\multicolumn{1}{ |l|}{PGD}
	   & 59.1    & 58.8
	& 47.9     & 85.7
	& 51.8     & 94.6
	& 66.3                  \\
	
    &\multicolumn{1}{|l|}{AutoAttack}
	  & 22.6    & 27.7
	& 34.7     & 21.8
	& 35.9     & 11.8
	& 25.8                  \\        
	\hline

	\multirow{4}{*}{CDAN +Trades}
    &\multicolumn{1}{ |l|}{None (clean)}
    &83.7 &81.1 &57.0 &96.2 &56.9 &98.0 &78.8\\
    
    &\multicolumn{1}{ |l|}{I-FSGM}
	& 52.4    & 49.2
	& 54.5     & 85.3
	& 51.4     & 90.6
	& 63.9                \\
	
    &\multicolumn{1}{ |l|}{PGD}
	  & 62.9    & 61.5
	& 49.0     & 83.1
	& 54.9     & 92.7
	& 64.7                \\
	
    &\multicolumn{1}{ |l|}{AutoAttack}
	& 36.1    & 34.3
	& 33.5     & 25.7
	& 40.5     & 26.2
	& 32.8                \\        
	\hline

	\multirow{4}{*}{CDAN + DART }
    &\multicolumn{1}{ |l|}{None (clean)}
    &\textbf{91.8 }&\textbf{94.1} &\textbf{72.7} &\textbf{98.6} &71.3&\textbf{99.2} &\textbf{88.0}\\
    
    &\multicolumn{1}{ |l|}{I-FSGM}
	& \textbf{91.6}   & \textbf{94.7}
	& \textbf{72.3}    & \textbf{98.2}
	& \textbf{69.3}     & \textbf{98.2}
	& \textbf{87.4}                \\
	&\multicolumn{1}{ |l|}{PGD}
	  & \textbf{91.2}    & \textbf{93.2}
	& \textbf{69.7}    & \textbf{97.7}
	& \textbf{69.5}     & \textbf{98.7}
	& \textbf{86.7}                \\
	&\multicolumn{1}{ |l|}{AutoAttack}
	         & \textbf{90.4}    & \textbf{92.7}
	& \textbf{67.8}    & \textbf{94.5}
	& \textbf{68.1}     & 95.8
	& \textbf{84.9}                \\        
	\hline
\end{tabular}}
\end{table}

\begin{table*}[t]
\centering
	\caption{Defense against different attacks on Office-Home. Baseline is CDAN. The pink-highlighted cells denote the upper bounds of clean accuracy.}
	\label{table:6}
 \setlength{\tabcolsep}{4pt}
	\resizebox{\linewidth}{!}
    {
		\begin{tabular}{p{10mm} c| cccccccccccc|c }
			\hline
		    \multicolumn{1}{ l}{Defense method}
	        & \multicolumn{1}{|l|}{Attack method}
			& {~Ar$\rightarrow$Cl~}
			& {Ar$\rightarrow$Pr~}
			& {Ar$\rightarrow$Rw~}
			& {Cl$\rightarrow$Ar~}
			& {Cl$\rightarrow$Pr~}
			& {Cl$\rightarrow$Rw~}
			& {Pr$\rightarrow$Ar~}
			& {Pr$\rightarrow$Cl~}
			& {Pr$\rightarrow$Rw~}
			& {Rw$\rightarrow$Ar~}
			& {Rw$\rightarrow$Cl~}
			& {Rw$\rightarrow$Pr~}
			& {Avg.} \\ \hline

\multirow{2}{*}{\textcolor{black}{ARTUDA~\cite{Lo_2022_ACCV}}}
    &\multicolumn{1}{ |l|}{\textcolor{black}{None (clean)}}
 &\textcolor{black}{47.5} &\textcolor{black}{34.9} &\textcolor{black}{40.4} &\textcolor{black}{21.6} &\textcolor{black}{43.2} &\textcolor{black}{40.4} &\textcolor{black}{27.7} &\textcolor{black}{46.8} &\textcolor{black}{49.5} &\textcolor{black}{32.2} &\textcolor{black}{54.9} &\textcolor{black}{68.1}&\textcolor{black}{42.3}  \\
&\multicolumn{1}{ |l|}{\textcolor{black}{AutoAttack}}
&  \textcolor{black}{32.3}     & \textcolor{black}{18.0}
& \textcolor{black}{21.2}     & \textcolor{black}{12.2}
& \textcolor{black}{27.1}    & \textcolor{black}{24.0}
& \textcolor{black}{9.8 }    & \textcolor{black}{37.4}
& \textcolor{black}{28.0 }    & \textcolor{black}{17.2}
& \textcolor{black}{43.7}     & \textcolor{black}{40.0}
& \textcolor{black}{25.9}         \\        \hline

\multirow{2}{*}{\textcolor{black}{RFA~\cite{awais2021adversarial}}}
    &\multicolumn{1}{ |l|}{\textcolor{black}{None (clean)}}
 &\textcolor{black}{47.5} &\textcolor{black}{53.8} &\textcolor{black}{63.0} &\textcolor{black}{43.6} &\textcolor{black}{59.4} &\textcolor{black}{57.2} &\textcolor{black}{42.3} &\textcolor{black}{47.6} &\textcolor{black}{64.1} &\textcolor{black}{54.9} &\textcolor{black}{55.6} &\textcolor{black}{72.8}&\textcolor{black}{55.1}  \\
&\multicolumn{1}{ |l|}{\textcolor{black}{AutoAttack}}
&  \textcolor{black}{31.6 }    & \textcolor{black}{29.1}
& \textcolor{black}{28.4}     & \textcolor{black}{16.3}
& \textcolor{black}{32.6}    & \textcolor{black}{25.8}
& \textcolor{black}{14.5}     & \textcolor{black}{28.3}
& \textcolor{black}{25.9}     & \textcolor{black}{19.0}
&  \textcolor{black}{33.3}     & \textcolor{black}{37.3}
& \textcolor{black}{26.8}         \\        \hline

\multirow{2}{*}{\textcolor{black}{SRoUDA~\cite{zhu2022srouda}}}
    &\multicolumn{1}{ |l|}{\textcolor{black}{None (clean)}}
 &\textcolor{black}{53.6} &\textbf{\textcolor{black}{75.2}} &\textbf{\textcolor{black}{78.6}} &\textcolor{black}{60.1} &\textcolor{black}{70.1} &\textbf{\textcolor{black}{70.1}} &\textbf{\textcolor{black}{61.6}} &\textcolor{black}{44.7} &\textbf{\textcolor{black}{79.4}} &\textbf{\textcolor{black}{72.6}} &\textcolor{black}{52.3} &\textbf{\textcolor{black}{83.6}}&\textcolor{black}{60.1}  \\
&\multicolumn{1}{ |l|}{\textcolor{black}{AutoAttack}}
& \textcolor{black}{46.6 }    & \textcolor{black}{66.6}
& \textcolor{black}{70.0 }    &\textcolor{black}{ 54.7}
& \textbf{\textcolor{black}{67.1}}    &  \textcolor{black}{62.7}
& \textbf{\textcolor{black}{58.5}}     & \textcolor{black}{41.5}
& \textbf{\textcolor{black}{71.1}}     & \textbf{\textcolor{black}{69.8}}
&  \textcolor{black}{46.3}     & \textbf{\textcolor{black}{80.4}}
& \textcolor{black}{54.7}         \\        \hline

\multirow{4}{*}{CDAN (None)}
    &\multicolumn{1}{ |l|}{None (clean)}
     &\cellcolor{hfxpink}53.3 &\cellcolor{hfxpink}71.9 &\cellcolor{hfxpink}78.4 &\cellcolor{hfxpink}61.4 &\cellcolor{hfxpink}72.1 &\cellcolor{hfxpink}72.2 &\cellcolor{hfxpink}62.4 &\cellcolor{hfxpink}55.3 &\cellcolor{hfxpink}80.5 &\cellcolor{hfxpink}74.8 &\cellcolor{hfxpink}60.4 &\cellcolor{hfxpink}84.2 &\cellcolor{hfxpink}68.9\\

		&\multicolumn{1}{ |l|}{I-FSGM}
& 29.7     & 39.0
& 41.7     & 34.7
& 43.1     & 45.4
& 23.7     & 15.9
& 32.0     & 35.2
& 28.9     & 44.5
& 34.5                \\
&\multicolumn{1}{ |l|}{PGD}
& 39.2     & 41.4
& 47.9     & 34.7
& 42.0     & 44.2
& 27.9     & 32.1
& 40.9     & 36.8
& 30.5     & 44.1
& 38.5                \\
&\multicolumn{1}{ |l|}{AutoAttack}
& 18.9     & 21.9
& 16.3     & 17.5
& 24.1     & 19.4
& 14.8     & 20.0
& 16.5     & 13.5
& 18.8     & 20.7
& 18.5                \\        \hline

\multirow{4}{*}{CDAN + AT}
    &\multicolumn{1}{ |l|}{None (clean)}
 &43.0 &55.9 &67.1 &49.0 &64.5 &61.3 &44.3 &44.4 &70.1 &64.2 &55.7 &78.6 &57.3   \\
&\multicolumn{1}{ |l|}{I-FSGM}
& 41.5     & 50.2
& 60.1     & 40.0
& 55.0     & 53.3
& 38.8     & 41.7
& 60.5     & 54.3
& 49.3     & 66.8
& 50.1          \\
&\multicolumn{1}{ |l|}{PGD}
& 46.3     & 51.6
& 63.0     & 40.1
& 56.9     & 54.0
& 42.3     & 44.5
& 63.0     & 54.4
& 50.5     & 60.4
& 51.4          \\
&\multicolumn{1}{ |l|}{AutoAttack}
& 29.5     & 33.3
& 27.3     & 24.3
& 32.3     & 21.5
& 22.4     & 28.2
& 21.3     & 25.2
& 30.8     & 35.5
& 27.6         \\        \hline

\multirow{4}{*}{CDAN + Trades}
    &\multicolumn{1}{ |l|}{None (clean)} 
    &49.8 &61.1 &69.5 &52.7 &68.9 &63.7 &56.4 &51.7 &71.7 &68.1 &60.9 &80.4 &62.9\\
    
&\multicolumn{1}{ |l|}{I-FSGM}
& 45.5     & 56.3
& 64.5     & 45.9
& 65.1     & 58.5
& 53.1     & 43.4
& 63.3     & 62.0
& 57.4     & 77.2
& 57.7          \\
&\multicolumn{1}{ |l|}{PGD}
& 46.6     & 54.5
& 60.3     & 47.9
& 59.2     & 55.7
& 47.1     & 45.2
& 64.1     & 61.3
& 51.9     & 67.6
& 54.9          \\
&\multicolumn{1}{ |l|}{AutoAttack}
& 32.4     & 42.6
& 35.9     & 31.8
& 34.4     & 27.1
& 20.7     & 29.6
& 28.0     & 29.6
& 27.9     & 36.2
& 31.4         \\        \hline
\multirow{4}{*}{CDAN + DART }
    &\multicolumn{1}{ |l|}{None (clean)} 
 &\textbf{53.8} & 66.2 & 71.8 & \textbf{60.2} & \textbf{70.8} & 68.1 & 60.0 & \textbf{54.5} & 79.0 & 71.9 & \textbf{60.6} & 82.2 & \textbf{66.6}   \\
&\multicolumn{1}{ |l|}{I-FSGM}
& \textbf{53.2}    & \textbf{67.1}
& \textbf{73.2}     & \textbf{59.6}
& \textbf{70.6}    & \textbf{70.4}
& \textbf{61.5}    & \textbf{54.4}
& \textbf{77.9}    & \textbf{70.3}
& \textbf{59.2}     & \textbf{81.7}
& \textbf{66.6}               \\
&\multicolumn{1}{ |l|}{PGD}
& \textbf{52.4}  & \textbf{69.5}
& \textbf{73.2}    & \textbf{57.6}
& \textbf{66.2}    & \textbf{68.3}
& \textbf{59.8}     & \textbf{53.1}
& \textbf{74.8}     & \textbf{70.1}
& \textbf{58.9}     & \textbf{77.3}
& \textbf{65.1}                \\
&\multicolumn{1}{ |l|}{AutoAttack}
& \textbf{51.8}    & 67.2
& \textbf{72.1}     & 56.9
& 65.6    & \textbf{63.9}
&56.9    & \textbf{51.0}
& 70.7   & 67.2
& \textbf{58.1}     & 79.4
& \textbf{63.4}               \\        \hline
	\end{tabular}}
\end{table*}

\begin{table*}[t]
	\caption{Accuracy (clean,  \%) and adversarial robustness (Adv.,  \%) for AutoAttack on DomainNet. Bold indicates that the proposed method performs better.}
 \vspace{-3mm}
\label{table:DomainNetaa}
\begin{center}
\setlength{\tabcolsep}{1.5pt}
\renewcommand{\arraystretch}{1}
	\resizebox{\linewidth}{!}{
		\begin{tabular}{p{0.5mm} c  c c c c c c c c c ||l |c c c c c c c c c c }
			\hline

\multicolumn{1}{ c|}{}
			& \multicolumn{2}{c}{clp}
			& \multicolumn{2}{c}{inf}
			& \multicolumn{2}{c}{pnt}
			& \multicolumn{2}{c}{rel}
			& \multicolumn{2}{c||}{skt}
			& \multicolumn{1}{c|}{}
			& \multicolumn{2}{c}{clp}
			& \multicolumn{2}{c}{inf}
			& \multicolumn{2}{c}{pnt}
			& \multicolumn{2}{c}{rel}
			& \multicolumn{2}{c}{skt}     \\
			
			\multicolumn{1}{ c|}{ARTUDA~\cite{Lo_2022_ACCV}}
			& clean    & Adv.     & clean    & Adv.
			& clean    & Adv.     & clean    & Adv.
			& clean    & Adv.    
			& \multicolumn{1}{ c|}{RFA~\cite{awais2021adversarial}}
			& clean    & Adv.     & clean    & Adv.
			& clean    & Adv.     & clean    & Adv.
			& clean    & Adv.          \\
			\hline

			\multicolumn{1}{c|}{clp}
			& -        & -
			& 12.6    & 3.8
			& 31.2    & 7.9
			& 44.9     & 16.5
			& 39.9     & 17.6  
            & \multicolumn{1}{ c|}{clp}
			& -        & -        & 9.4     & 3.5
			& 22.3     & 7.6     & 37.5     & 17.7
			& 27.4     & 12.2    
            \\
			
			\multicolumn{1}{ c|}{inf}
			&31.9     & 12.1     & -        & -
			& 20.4     & 2.6     & 23.8      & 5.5
			& 34.6     & 5.9   
			& \multicolumn{1}{c|}{inf}
			& 15.0     & 8.8
			& -        & -
			& 14.8     & 4.8
			& 19.2    & 8.8
			& 11.4     & 5.1
             \\
			
			\multicolumn{1}{ c|}{pnt}
			& 42.9     & 20.6     & 14.4     & 3.2
			& -        & -        & 49.0      & 16.6
			& 34.0     & 13.1
			& \multicolumn{1}{c|}{pnt}
			& 26.3     & 14.9
			& 9.5     & 3.1
			& -        & -
			& 37.5    & 17.0
			& 19.7    & 8.6
			   \\
			
			\multicolumn{1}{ c|}{rel}
			& 52.9     & 28.1     & 16.9   & 4.2
			& 42.0     & 12.2     & -      & -
			& 36.1        & 14.9       
			& \multicolumn{1}{c|}{rel}
			& 36.7     & 19.3
			& 12.4    & 3.0
			& 32.0    & 8.3
			& -     &-
			& 22.7        & 8.1
		   \\
			
			\multicolumn{1}{ c|}{skt}
			& 54.2    & 30.6     &13.1     & 3.9
			& 34.9    & 7.8    & 41.4     &13.5
	      & -        & -
			& \multicolumn{1}{c|}{skt}
			& 33.9    & 19.9     &9.5     & 3.5
			& 27.3    & 7.8    &30.4     &13.7
		  & -        & -       \\
			\hline
			\hline

\multicolumn{1}{ c|}{}
			& \multicolumn{2}{c}{clp}
			& \multicolumn{2}{c}{inf}
			& \multicolumn{2}{c}{pnt}
			& \multicolumn{2}{c}{rel}
			& \multicolumn{2}{c||}{skt}
			& \multicolumn{1}{c|}{DART}
			& \multicolumn{2}{c}{clp}
			& \multicolumn{2}{c}{inf}
			& \multicolumn{2}{c}{pnt}
			& \multicolumn{2}{c}{rel}
			& \multicolumn{2}{c}{skt}     \\
			
			\multicolumn{1}{ c|}{SRoUDA~\cite{zhu2022srouda}}
			& clean    & Adv.     & clean    & Adv.
			& clean    & Adv.     & clean    & Adv.
			& clean    & Adv.    
			& \multicolumn{1}{ c|}{(ours)}
			& clean    & Adv.     & clean    & Adv.
			& clean    & Adv.     & clean    & Adv.
			& clean    & Adv.          \\
			\hline

			\multicolumn{1}{c|}{clp}
			& -        & -
			& 9.3    & 5.7
			& 24.3    & 15.1
			& 42.9     & 28.4
			& 32.4     & 23.8  
            & \multicolumn{1}{ c|}{clp}
			& -        & -        & \textbf{19.4}     & \textbf{8.3}
			& \textbf{35.3}     & \textbf{16.6}     & \textbf{53.3}     & 25.6
			& \textbf{41.5}     & 22.1    
            \\
			
			\multicolumn{1}{ c|}{inf}
			&26.0     & 14.2     & -        & -
			& 24.3     & 13.9     & 36.3      & 22.1
			& 22.1     & 13.5   
			& \multicolumn{1}{c|}{inf}
			& \textbf{33.8}    & \textbf{25.3}
			& -        & -
			& \textbf{32.7}     & 8.5
			&\textbf{ 49.3}    & 8.6
			& 22.8     & 10.1
             \\
			
			\multicolumn{1}{ c|}{pnt}
			& 36.7     & 28.3     & 10.2     & 5.9
			& -        & -        & 45.7      & 30.0
			& 31.0     & 19.1
			& \multicolumn{1}{c|}{pnt}
			& 41.1     & 25.4
			& \textbf{17.9 }   &\textbf{ 6.9}
			& -        & -
			& \textbf{57.9}    & 19.2
			& \textbf{36.6}    & \textbf{19.8}
			   \\
			
			\multicolumn{1}{ c|}{rel}
			& 44.3     & 34.3     & 10.1   & 5.6
			& 40.3     & 23.3     & -      & -
			& 29.1        & 18.6      
			& \multicolumn{1}{c|}{rel}
			& 50.4     & \textbf{36.2}
			& \textbf{20.6}    & 4.2
			& \textbf{51.8}    & \textbf{28.7}
			& -     &-
			& 33.1        &\textbf{ 18.7}
		   \\
			
			\multicolumn{1}{ c|}{skt}
			& 46.6    & 30.7     &8.5     & 5.3
			& 31.1    & 15.1    & 43.0     &22.3
	      & -        & -
			& \multicolumn{1}{c|}{skt}
			& \textbf{54.6}    & \textbf{40.8}     &13.0     & \textbf{5.3}
			& \textbf{44.3}    & \textbf{15.2}    &\textbf{52.0}     &14.6
		  & -        & -       \\
			\hline

	\end{tabular}}
\end{center}
\end{table*}

\begin{table*}[t]
	\caption{Comparative experiments with adversarial samples generated using pseudo labels and ground-truth labels of target,  respectively. Note that in standard UDA setting,  the ground-truth labels are agnostic,  which are just used here to explore the rationality using the pseudo labels.}
	\label{table:7}
	\begin{center}
 \vspace{-5mm}
 \setlength{\tabcolsep}{5pt}
\renewcommand{\arraystretch}{1.2}
		\resizebox{\linewidth}{!}
        {
			\begin{tabular}{ c|c|cccccccccccc |c }
				\hline
				{Label}&sample
				& {Ar$\rightarrow$Cl~}
				& {Ar$\rightarrow$Pr~}
				& {Ar$\rightarrow$Rw}~
				& {Cl$\rightarrow$Ar}~
				& {Cl$\rightarrow$Pr}~
				& {Cl$\rightarrow$Rw}~
				& {Pr$\rightarrow$Ar}~
				& {Pr$\rightarrow$Cl}~
				& {Pr$\rightarrow$Rw}~
				& {Rw$\rightarrow$Ar}~
				& {Rw$\rightarrow$Cl}~
				& {Rw$\rightarrow$Pr}~
				& {Avg.} \\
                
				\hline
				
				\multirow{2}{*}{pseudo labels} &{clean}
&\textbf{53.8} & 66.2 & \textbf{71.8} & 60.2 & 70.8 & 68.1 & 60.0 & 54.5 & \textbf{79.0} & \textbf{71.9} & \textbf{60.6} & \textbf{82.2} & 66.6 \\
&{Adv.}&53.2 & \textbf{67.1} & 73.2 & 59.6 & 70.6 & \textbf{70.4} & \textbf{61.5} & 54.4 & \textbf{77.9} & 70.3 & \textbf{59.2} & \textbf{81.7} & 66.6\\
				\hline
				\multirow{2}{*}{ground-truth} &{clean}
&53.5 & \textbf{67.5} & 71.3 & \textbf{60.5} & \textbf{71.1} & \textbf{69.2} & \textbf{61.3} & \textbf{55.4} & \textbf{79.0} & 71.5 & 60.3 & 81.9 & \textbf{66.9}       \\
&{Adv.}&\textbf{53.6} & 67.0 & \textbf{74.7} & \textbf{61.4} & \textbf{70.9} & 69.3 & 61.0 & \textbf{54.8} & 77.2 & \textbf{70.8} & 59.0 & \textbf{81.3} & \textbf{66.8}\\
				\hline
		\end{tabular}}
	\end{center}
\end{table*}

\begin{figure*}[t]
	\centering
	\includegraphics[width=1\textwidth]{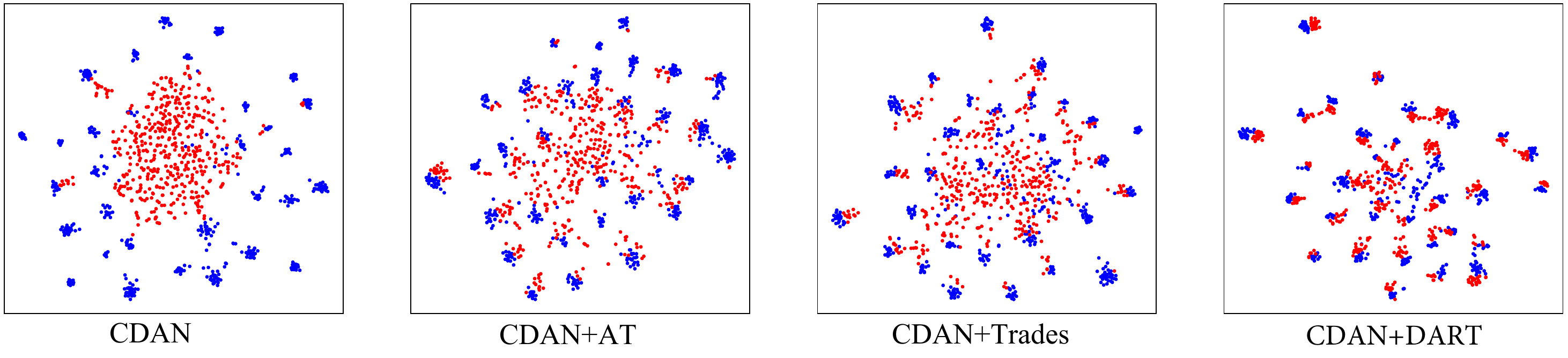}
	\caption{Visualization by t-SNE on A$\rightarrow$D,  where red and blue points indicate the clean and adversarial samples,  respectively.}
	\label{fig:4}
\end{figure*}

\textbf{Results on Office-31.} Table \ref{table:Office-31} shows the quantitative results using different baselines. As expected,  our DART paradigm achieves noticeable performance gains, and surpasses  +AT and +Trades based robust UDA models and previous robust UDA models by a large margin in both clean accuracy and adversarial robustness. In particular, our DART improves the accuracy of the adversarial samples while maintaining the accuracy of the clean samples (slightly degrades less than 1\%).

\textbf{Results on Office-Home.} Table \ref{table:Office-home} shows the robustness of +AT and +Trades is improved compared to the non-robust UDA model (i.e.,  ``None''),  but the accuracy on clean samples drops by more than 15\%,  which verifies the disadvantage of them. In contrast,  compared with ``None'',  the accuracy of the model trained by our DART drops slightly,  which is close to the upper bounds (i.e.,  ``None''). Overall, the proposed DART with different UDA baselines shows the best robustness.

\textbf{Results on Amazon Reviews.} Table \ref{table:Office-home} shows the performance on the pseudo-labeling based method, i.e., PL-Mix. Compared to +AT, the proposed method significantly improves the performance on both clean and adversarial examples. This demonstrates the universality and compatibility of DART for improving the robustness of different UDA methods.

\textbf{Results on VisDA-2017.} As shown in Table \ref{table:VisDA-2017},  our DART achieves the optimal accuracy. Besides, the accuracy of adversarial samples via +DART even aligns with the accuracy of the clean samples. These results reveal that the proposed DART achieves the best adversarial robustness and simultaneously overcomes the accuracy degradation of clean samples. 

\textbf{Results on DomainNet.} DART achieves a very high average accuracy on the most challenging DomainNet dataset,  as shown in Table \ref{table:DomainNet}. Incredibly,  DART surpasses other methods in all 30 tasks,  which demonstrates the strong ability of DART to alleviate large domain gap and improve noise robustness. 

To sum up, the results on multiple benchmarks clearly indicate the proposed URDA paradigm with a new generalization bound is superior to the UDA+VAT paradigm,  and the DART algorithm induced from the proposed URDA theory performs the best among existing robust UDA models.

\section{Analysis and Discussion}
\textbf{Generalizability to Attacks.}  To further demonstrate the generalization ability of our proposed DART against different attacks,  we conduct the experiment on three powerful attacks,  including I-FGSM \cite{kurakin2018adversarial},  PGD \cite{madry2017towards} and AutoAttack \cite{croce2020reliable} as shown in Table \ref{table:5}, \ref{table:6}, and \ref{table:DomainNetaa}. Our proposed DART exhibits good robustness under three different attacks. Even under the powerful AutoAttack,  DART can still maintain good performance,  while the robustness of other three methods has been significantly decreased. This indicates DART has strong generalization ability against different attacks.

 \textbf{Impact of target pseudo labels.} In experiments,  we use pseudo labels to generate adversarial samples. In the field of adversarial defense,  adversarial samples are usually generated using the ground-truth labels based on targeted or non-targeted attacks. However,  in UDA tasks,  the target domain does not have ground-truth labels. Considering that the goal of adversarial samples is to cheat the model by feeding wrong class labels,  although the target pseudo labels often contain noise,  the pseudo target labels impose little impact because they still resort to generating fake samples of wrong classes. In order to prove that pseudo labels have little impact on the robustness,  we conduct some exploratory experiments by exploiting the ground-truth target labels to generate adversarial samples for comparison. As shown in Table \ref{table:7},  the results indicate that the adversarial samples generated by target pseudo labels have almost no impact on the adversarial robustness of the model. This confirms that it is feasible to generate adversarial target samples by using target pseudo labels.
	
\textbf{Feature Visualization.}
The t-SNE~\cite{van2008visualizing} visualizations of features learned by CDAN (None),  +AT,  +Trades and +DART on the A $\rightarrow$ D task are presented in Fig. \ref{fig:4}. It is evident that the features of clean samples from the same class generated by CDAN are clustered together and a distinct classification boundary exists between classes. However,  the features of adversarial samples from different classes appear to be confused and misclassified. The vulnerability of CDAN,  a vanilla domain adaptation model,  is exhibited. The improved CDAN with robust operators,  i.e. +AT and +Trades alleviate this problem to some extent,  but still not ideal. This is due to the entanglement challenge between transfer training and robust training in UDA+VAT paradigm is not catched and resolved. In contrast,  with +DART,  the features of adversarial samples from the same class are not only well-clustered but also mostly situated near the homogeneous clean samples,  indicating the robustness of URDA paradigm against adversarial attacks. 

\begin{figure}[t]
	\centering
	\includegraphics[width=0.45\textwidth]{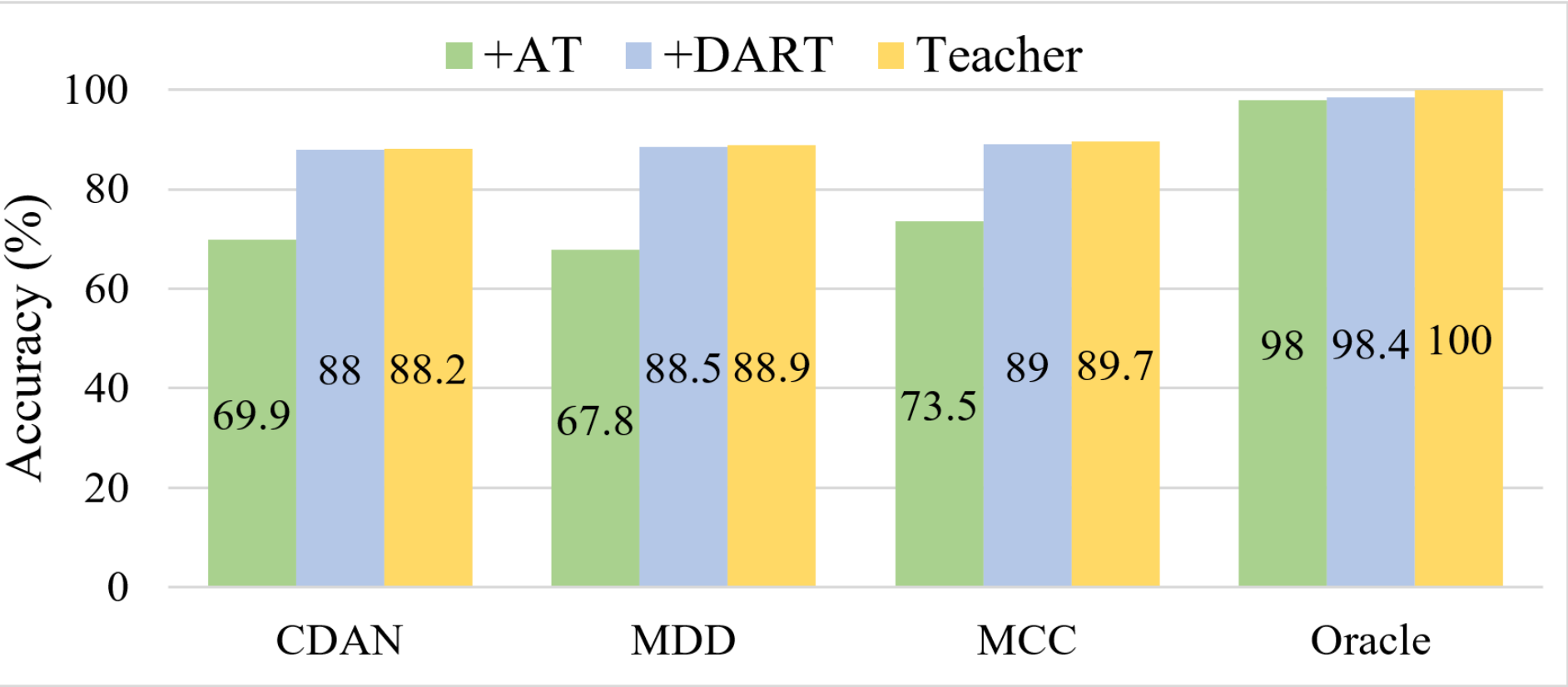}
	\caption{Accuracy of clean target samples for different robust UDA model and its upper bound (teacher) based on different ideal target classifiers $h_{t}^{*}$ including CDAN, MDD, MCC and Oracle, respectively. Note that Oracle means the ideal target classifier is trained using the ground-truth target labels, which is actually unavailable. Office-31 dataset is experimented.}
	\label{fig:5}
\end{figure}



\begin{figure}[t]
	\centering
	\includegraphics[width=0.48\textwidth]{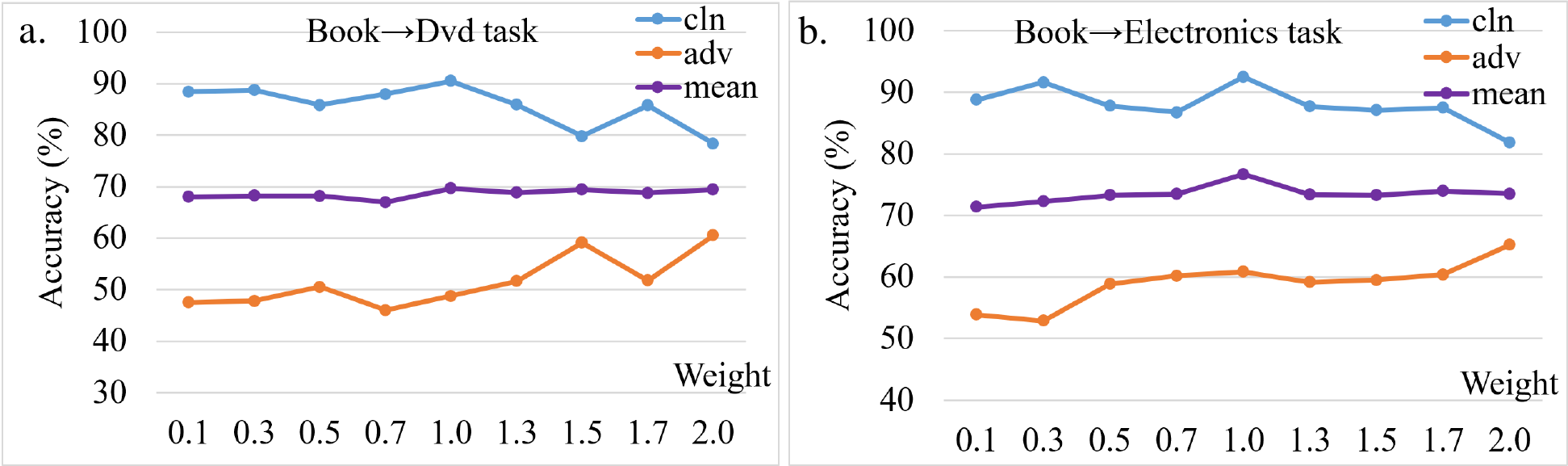}
	\caption{Weight analysis of MSE loss in the proposed method on the Amazon Reviews dataset. 
    }
	\label{weight}
\end{figure}

\textbf{Impact of the ideal target classifier $h_{t}^{*}$.} In the proposed theoretical upper bound,  as shown in Definition \ref{definition1},  we defined an ideal target classifier in Eq. \ref{eqtargetclassifier},  in order to facilitate the foundation of the robust domain adaptation theory for the DART method. In the actual implementation process,  we replace the ideal target classifier with the existing UDA method (e.g.,  CDAN,  MDD and MCC),  and the performance of this UDA method determines the upper limit of our method. Generally,  the higher the target accuracy of the vanilla UDA model (a.k.a. teacher),  the better the effect of the DART method,  as is shown in Fig. \ref{fig:5}. From the perspective of robustness,  treating the vanilla UDA as the ideal target classifier is rational,  because DART is designed as a firewall to protect the UDA model from being attacked without degrading clean accuracy,  rather than upgrading the clean accuracy of UDA. To further confirm the perspective,  we train an ideal target classifier with ground-truth labels of the target domain (note that the target labels are only used here for exploratory experiments,  which are actually unknown). In Fig. \ref{fig:5},  under the ideal target classifier,  the clean target accuracy achieves 100\%,  and the robustness is also improved. Additionally,  the proposed DART succeeds to improve the robustness of vanilla UDA model towards the upper bound (e.g., $88.0\to 88.2$,  $88.5\to 88.9$ and $89.0\to 89.7$ for CDAN,  MDD and MCC,  respectively).

\begin{figure}[t]
	\centering
	\includegraphics[width=0.48\textwidth]{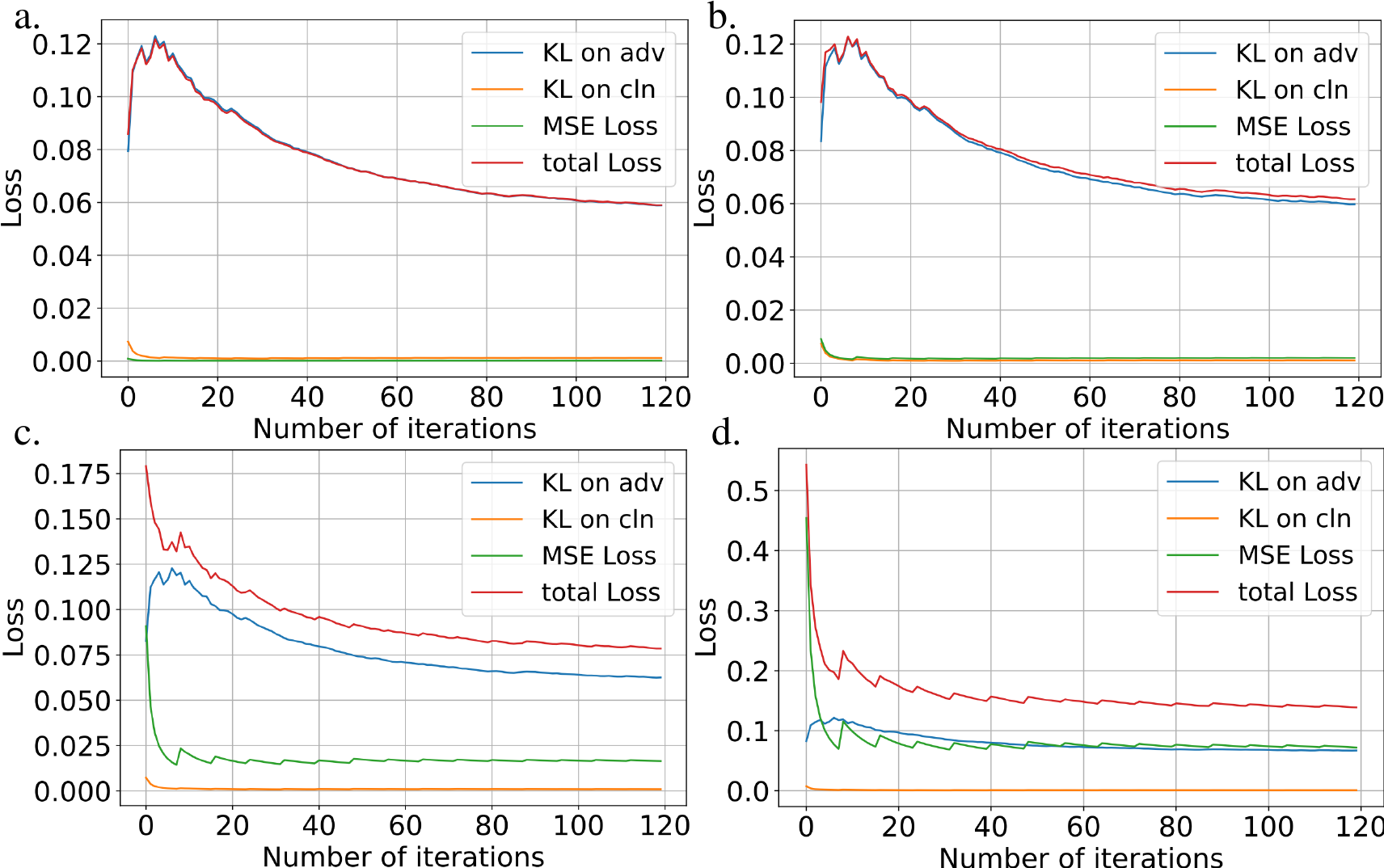}
	\caption{Convergence curve during training of the MSE loss with different weights. a., b., c., d. represent the curves when the weights are 0.01, 0.1, 1 and 5, respectively.}
	\label{curve}
\end{figure}

\textbf{Impact of MSE loss weight and optimization stability.} In Step 2, the MSE loss is a key component used for feature distillation, so it is important to investigate the impact of its weight during training. As shown in Fig.~\ref{weight}, based on the pseudo-label based UDA method i.e. PL-Mix, we conduct quantitative experiments on the Amazon Reviews  dataset. As can be seen, the performance on both clean and adversarial examples remains consistent across a range of loss weights [0.1, 2], with optimal results achieved when the weight is set to 1. Fig.~\ref{curve} shows the convergence curves of different losses at different weight during training. Across all settings, the three loss terms converge smoothly, and the total loss decreases steadily. Although larger MSE weights slightly slow down the decay of adversarial consistency, no divergence or oscillation is observed. These results indicate that the proposed disentangled training paradigm has a good optimization stability.

\begin{table}[t]
    \caption{Ablation study for losses. Baseline is CDAN.}
    \centering
    \footnotesize  
    \begin{tabular}{@{}lcc@{\hspace{0.5em}}cc@{\hspace{0.5em}}cc@{}}
        \hline
        \multirow{2}{*}{Method} & \multicolumn{2}{c}{W$\rightarrow$A} & \multicolumn{2}{c}{A$\rightarrow$W} & \multicolumn{2}{c}{Avg.} \\
        & Clean & Adv. & Clean & Adv. & Clean & Adv. \\
        \hline
        w/o. $MSE(\cdot)$  & 63.9 & 48.2 & 89.2 & 82.8 & 76.6 & 65.5 \\
        w/o. $KL(\cdot)$   & 65.1 & 48.6 & 87.9 & 76.4 & 76.5 & 62.5 \\
        w/o. $\mathcal{L}_{\text{T}}$ & 50.3 & 36.1 & 82.6 & 72.6 & 66.4 & 54.3 \\
        w/o. $\mathcal{L}_{\widetilde{T}}$ & 68.5 & 47.3 & 89.2 & 22.1 & 78.9 & 34.7 \\
        \hline
        DART & \textbf{71.3} & \textbf{69.3} & \textbf{94.1} & \textbf{94.7} & \textbf{82.7} & \textbf{82.0} \\
        \hline
    \end{tabular}
    \label{table:ablation}
\end{table}

\textbf{Ablation Study.}
To investigate the impact of each loss function,  we conduct ablation experiments on A$\rightarrow$W and W$\rightarrow$A tasks. As shown in Table \ref{table:ablation},  when the model lacks the $MSE(\cdot)$ loss in Eq. (\ref{distll_clean}),  the accuracies of both clean samples and adversarial samples decrease. Without $KL(\cdot)$ loss,  the robustness of the model decreases more severely. This reveals that the model distillation in prediction level has a greater impact on the adversarial robustness. Besides,  when both $MSE(\cdot)$ and $KL(\cdot)$ are not used (i.e.,  the $\mathcal{L}_T$ loss in Eq. (\ref{distll_clean}) is removed),  the accuracy of the model to clean samples cannot be guaranteed,  and the robustness brought by $\mathcal{L}_{\widetilde{T}}$ is also limited. When $\mathcal{L}_{\widetilde{T}}$ is removed,  it is clear that the adversarial robustness cannot be guaranteed,  due to the absence of robustness training. In short,  $\mathcal{L}_{\widetilde{T}}$ is the key to improving adversarial robustness of the model,  while $\mathcal{L}_T$ is the key to keeping the accuracy of clean data.

\section{Conclusion,  Limitation and Outlook}
This paper explores to establish a new URDA paradigm,  theory and algorithm for protecting UDA against adversarial samples without scarifying accuracy of benign samples. After delving deeper into the challenges of adversarial training in UDA,  we uncover that the undermined problem preventing robustness training is the entanglement of transfer training and adversarial training in UDA+VAT paradigm. Building on this insight,  URDA paradigm (i.e.,  unsupervised robust domain adaptation) is arise. Evolved from the classical UDA theory,  we then derive a new generalization bound theory of URDA under noises for the first time by effectively disentangling the adversarial and transfer training.
Given the theoretical perspective,  we further propose a simple but effective training algorithm for URDA,  i.e.,  Disentangled Adversarial Robustness Training (DART). DART is a two-step robustness training algorithm,  including an \textit{off-line transfer step} by pre-training a traditional UDA model and an \textit{on-the-fly robustification step} by post-training a robust model. Exhaustive experiments validate that the proposed DART can improve the adversarial robustness to perturbations while retaining the accuracy of benign samples.
Crucially,  DART is independent of the traditional UDA training,  which facilitates the robustification post-training in practical application by only performing Step 2 on the fly while the pre-trained UDA model is frozen.

Although URDA paradigm is theoretically complete,  it is built under an assumption that the adversarial attacks are known,  which is commonly unknown in reality. This is always a tricky problem in adversarial attack and defense community,  and not a key focus of this paper,  since this paper just aims to establish an unsupervised robust domain adaptation framework under attacks. Therefore,  this can be an open issue for universal robustness to agnostic attacks,  and deserves much endeavor in future work.


\noindent \textbf{Acknowledgement.} This work was partially supported by National Natural Science Fund of China under Grants 92570110 and 62271090, Chongqing Natural Science Fund under Grant CSTB2024NSCQ-JQX0038, National Key R\&D Program of China under Grant 2021YFB3100800 and National Youth Talent Project. 


\noindent \textbf{Data Availability.} The Office-31 dataset involved in Table~\ref{table:Office-31} can be publicly accessed at \url{https://faculty.cc.gatech.edu/~judy/domainadapt/}. The Office-Home and Amazon Reviews datasets involved in Table~\ref{table:Office-home} can be publicly accessed at \url{https://www.hemanthdv.org/officeHomeDataset.html} and \url{https://www.cs.jhu.edu/~mdredze/datasets/sentiment/domain_sentiment_data.tar.gz}, respectively. The VisDA-2017 Dataset used in Table~\ref{table:VisDA-2017} can be obtained at \url{https://ai.bu.edu/visda-2017/}. The DomainNet dataset used in Table~\ref{table:DomainNet} are from \url{https://ai.bu.edu/M3SDA/}.

\bibliographystyle{unsrt}
\bibliography{sn-bibliography}

\end{document}